\documentclass[10pt,twocolumn,letterpaper]{article}

\usepackage{cvpr}
\usepackage{times}
\usepackage{epsfig}
\usepackage{graphicx}
\usepackage{amsmath}
\usepackage{amssymb}
\usepackage{algorithm}
\usepackage{algpseudocode}
\usepackage{rotating}
\usepackage[font={small}]{caption}
\usepackage{subcaption}
\usepackage{enumitem}
\usepackage{afterpage}

\graphicspath{{figures/}{figures/teaser/}}

\usepackage[pagebackref=true,breaklinks=true,letterpaper=true,colorlinks,bookmarks=false]{hyperref}

\cvprfinalcopy 

\def\cvprPaperID{0001} 

\ifcvprfinal\fi
\begin{document}

\title{Bayesian SegNet: Model Uncertainty in Deep Convolutional Encoder-Decoder Architectures for Scene Understanding}

\author{Alex Kendall \and Vijay Badrinarayanan\\
University of Cambridge\\
{\tt\small agk34, vb292, rc10001 @cam.ac.uk}
\and Roberto Cipolla
}

\maketitle

\begin{abstract}

We present a deep learning framework for probabilistic pixel-wise semantic segmentation, which we term Bayesian SegNet. Semantic segmentation is an important tool for visual scene understanding and a meaningful measure of uncertainty is essential for decision making. Our contribution is a practical system which is able to predict pixel-wise class labels with a measure of model uncertainty. We achieve this by Monte Carlo sampling with dropout at test time to generate a posterior distribution of pixel class labels. In addition, we show that modelling uncertainty improves segmentation performance by 2-3\% across a number of state of the art architectures such as SegNet, FCN and Dilation Network, with no additional parametrisation. We also observe a significant improvement in performance for smaller datasets where modelling uncertainty is more effective. We benchmark Bayesian SegNet on the indoor SUN Scene Understanding and outdoor CamVid driving scenes datasets.

\end{abstract}

\section{Introduction}

\begin{figure}[t]
\begin{center}
Input Images
\resizebox{\linewidth}{!}{
\makebox[\linewidth][c]{
\includegraphics[height=0.33\linewidth]{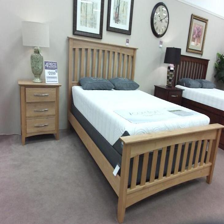}
\includegraphics[height=0.33\linewidth]{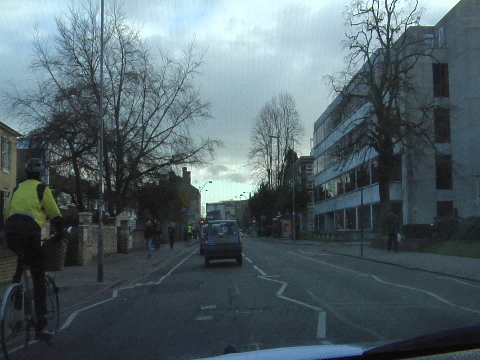}
\includegraphics[height=0.33\linewidth]{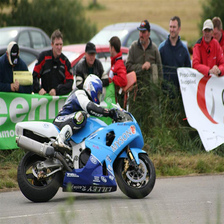}
}}
Bayesian SegNet Segmentation Output
\resizebox{\linewidth}{!}{
\makebox[\linewidth][c]{
\includegraphics[height=0.33\linewidth]{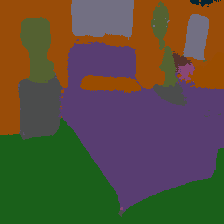}
\includegraphics[height=0.33\linewidth]{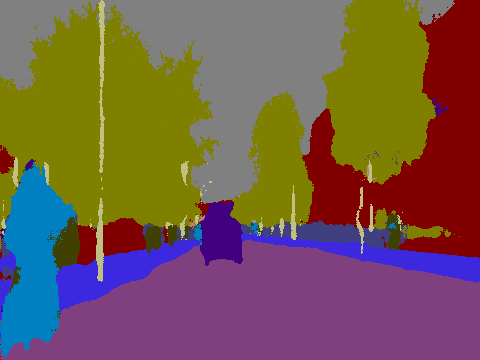}
\includegraphics[height=0.33\linewidth]{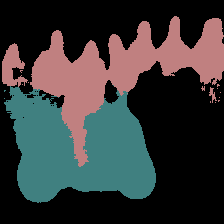}
}}
Bayesian SegNet Model Uncertainty Output
\resizebox{\linewidth}{!}{
\makebox[\linewidth][c]{
\includegraphics[height=0.33\linewidth]{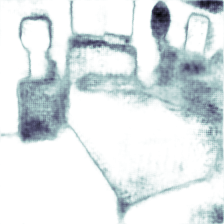}
\includegraphics[height=0.33\linewidth]{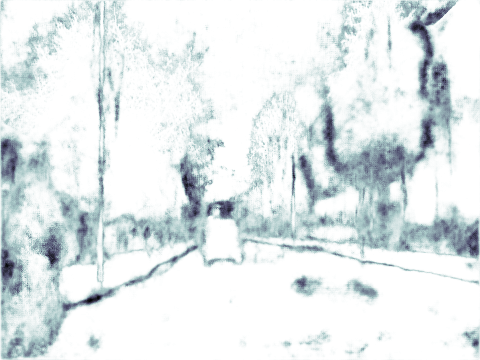}
\includegraphics[height=0.33\linewidth]{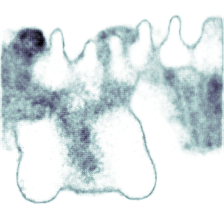}
}}
\end{center}
   \caption{\textbf{Bayesian SegNet. } These examples show the performance of Bayesian SegNet on popular segmentation and scene understanding benchmarks: SUN \cite{song2015sun} (left), CamVid \cite{brostow2009semantic} (center column) and Pascal VOC \cite{everingham2010pascal} (right). The system takes an RGB image as input (top), and outputs a semantic segmentation (middle row) and model uncertainty estimate, averaged across all classes (bottom row). We observe higher model uncertainty at object boundaries and with visually difficult objects. An online demo and source code can be found on our project webpage \href{mi.eng.cam.ac.uk/projects/segnet/}{mi.eng.cam.ac.uk/projects/segnet/}}
\label{fig:teaser}
\end{figure}

Semantic segmentation requires an understanding of an image at a pixel level and is an important tool for scene understanding. It is a difficult problem as scenes often vary significantly in pose and appearance. However it is an important problem as it can be used to infer scene geometry and object support relationships. This has wide ranging applications from robotic interaction to autonomous driving.

Previous approaches to scene understanding used low level visual features \cite{shotton2009textonboost}. We are now seeing the emergence of machine learning techniques for this problem \cite{shotton2013real,long2014fully}. In particular deep learning \cite{long2014fully} has set the benchmark on many popular datasets \cite{everingham2010pascal,couprie2013indoor}. However none of these deep learning methods produce a probabilistic segmentation with a measure of model uncertainty.

Uncertainty should be a natural part of any predictive system's output. Knowing the confidence with which we can trust the semantic segmentation output is important for decision making. For instance, a system on an autonomous vehicle may segment an object as a pedestrian. But it is desirable to know the model uncertainty with respect to other classes such as street sign or cyclist as this can have a strong effect on behavioural decisions. Uncertainty is also immediately useful for other applications such as active learning \cite{cohn1996active}, semi-supervised learning, or label propagation \cite{badrinarayanan2010label}.

The main contribution of this paper is extending deep convolutional encoder-decoder neural network architectures \cite{badrinarayanan2015segnet} to Bayesian convolutional neural networks which can produce a probabilistic segmentation output \cite{Gal2015DropoutB}. In Section \ref{sec:bayesian_segnet} we propose Bayesian SegNet, a probabilistic deep convolutional neural network framework for pixel-wise semantic segmentation.  We use dropout at test time which allows us to approximate the posterior distribution by sampling from the Bernoulli distribution across the network's weights. This is achieved with no additional parameterisation.

In Section \ref{sec:results}, we demonstrate that Bayesian SegNet sets the best performing benchmark on prominent scene understanding datasets, CamVid Road Scenes \cite{brostow2009semantic} and SUN RGB-D Indoor Scene Understanding \cite{song2015sun}. In particular, we find a larger performance improvement on smaller datasets such as CamVid where the Bayesian Neural Network is able to cope with the additional uncertainty from a smaller amount of data.

Moreover, we show in section \ref{sec:generalapplicable} that this technique is broadly applicable across a number of state of the art architectures and achieves a 2-3\% improvement in segmenation accuracy when applied to SegNet \cite{badrinarayanan2015segnet}, FCN \cite{long2014fully} and Dilation Network \cite{YuKoltun2016}.

Finally in Section \ref{sec:uncertainty} we demonstrate the effectiveness of model uncertainty. We show this measure can be used to understand with what confidence we can trust image segmentations. We also explore what factors contribute to Bayesian SegNet making an uncertain prediction.

\section{Related Work}

Semantic pixel labelling was initially approached with TextonBoost \cite{shotton2009textonboost}, TextonForest \cite{shotton2008semantic} and Random Forest Based Classifiers \cite{shotton2013real}. We are now seeing the emergence of deep learning architectures for pixel wise segmentation, following its success in object recognition for a whole image \cite{krizhevsky2012imagenet}. Architectures such as SegNet \cite{badrinarayanan2015segnet} Fully Convolutional Networks (FCN) \cite{long2014fully} and Dilation Network \cite{YuKoltun2016} have been proposed, which we refer to as the \textit{core segmentation engine}. FCN is trained using stochastic gradient descent with a stage-wise training scheme. SegNet was the first architecture proposed that can be trained end-to-end in one step, due to its lower parameterisation.

We have also seen methods which improve on these core segmentation engine architectures by adding post processing tools. HyperColumn \cite{hariharan2014hypercolumns} and DeConvNet \cite{noh2015learning} use region proposals to bootstrap their \textit{core segmentation engine}. DeepLab \cite{chen2014semantic} post-processes with conditional random fields (CRFs) and CRF-RNN \cite{zheng2015conditional} use recurrent neural networks. These methods improve performance by smoothing the output and ensuring label consistency. However none of these proposed segmentation methods generate a probabilistic output with a measure of model uncertainty.

Neural networks which model uncertainty are known as Bayesian neural networks \cite{denker1991transforming,mackay1992practical}. They offer a probabilistic interpretation of deep learning models by inferring distributions over the networks’ weights. They are often computationally very expensive, increasing the number of model parameters without increasing model capacity significantly. Performing inference in Bayesian neural networks is a difficult task, and approximations to the model posterior are often used, such as variational inference \cite{graves2011practical}.

On the other hand, the already significant parameterization of convolutional network architectures leaves them particularly susceptible to over-fitting without large amounts of training data. A technique known as \textit{dropout} is commonly used as a regularizer in convolutional neural networks to prevent overfitting and co-adaption of features \cite{srivastava2014dropout}. During training with stochastic gradient descent, \textit{dropout} randomly removes units within a network. By doing this it samples from a number of thinned networks with reduced width. At test time, standard dropout approximates the effect of averaging the predictions of all these thinnned networks by using the weights of the unthinned network. This is referred to as \textit{weight averaging}.

Gal and Ghahramani \cite{Gal2015DropoutB} have cast dropout as approximate Bayesian inference over the network's weights. \cite{Gal2015Bayesian} shows that dropout can be used at test time to impose a Bernoulli distribution over the convolutional net filter's weights, without requiring any additional model parameters. This is achieved by sampling the network with randomly dropped out units at test time. We can consider these as Monte Carlo samples obtained from the posterior distribution over models. This technique has seen success in modelling uncertainty for camera relocalisation \cite{kendall2015modelling}. Here we apply it to pixel-wise semantic segmentation.

\begin{figure*}[t]
	\captionsetup[sub]{font=small,labelfont={sf,sf}}
	\captionsetup[subfigure]{justification=centering}
	\begin{center}
		\resizebox{\linewidth}{!}{
		\begin{subfigure}[t]{0.2\linewidth}
			\includegraphics[width=\linewidth,trim={4cm 3cm 0cm 2cm},clip]{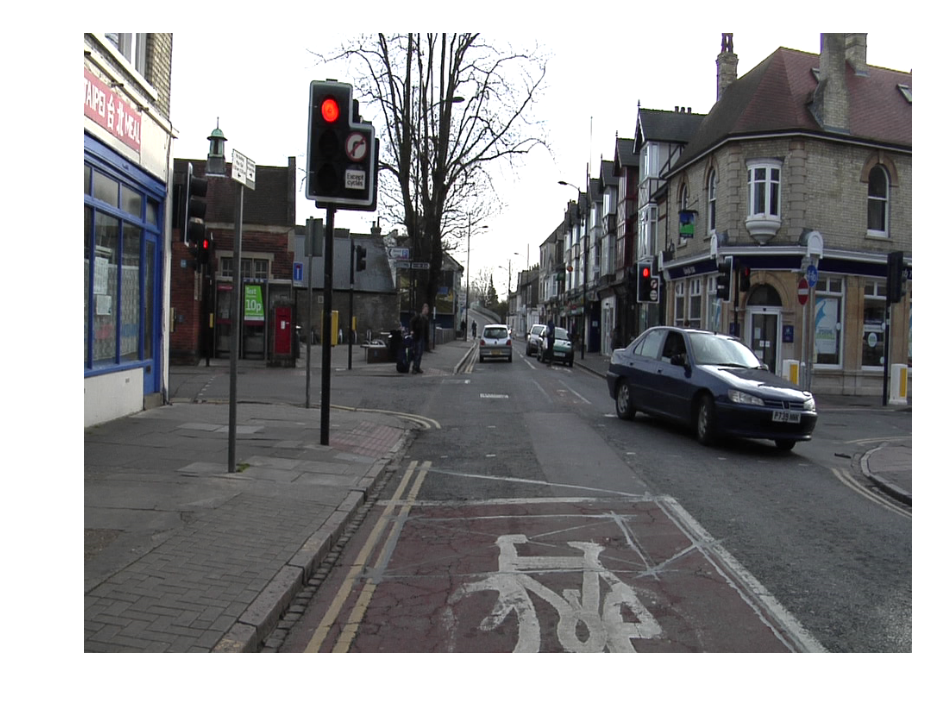}
			\caption{Input Image}
		\end{subfigure}
		\begin{subfigure}[t]{0.2\linewidth}
			\includegraphics[width=\linewidth,trim={4cm 3cm 0cm 2cm},clip]{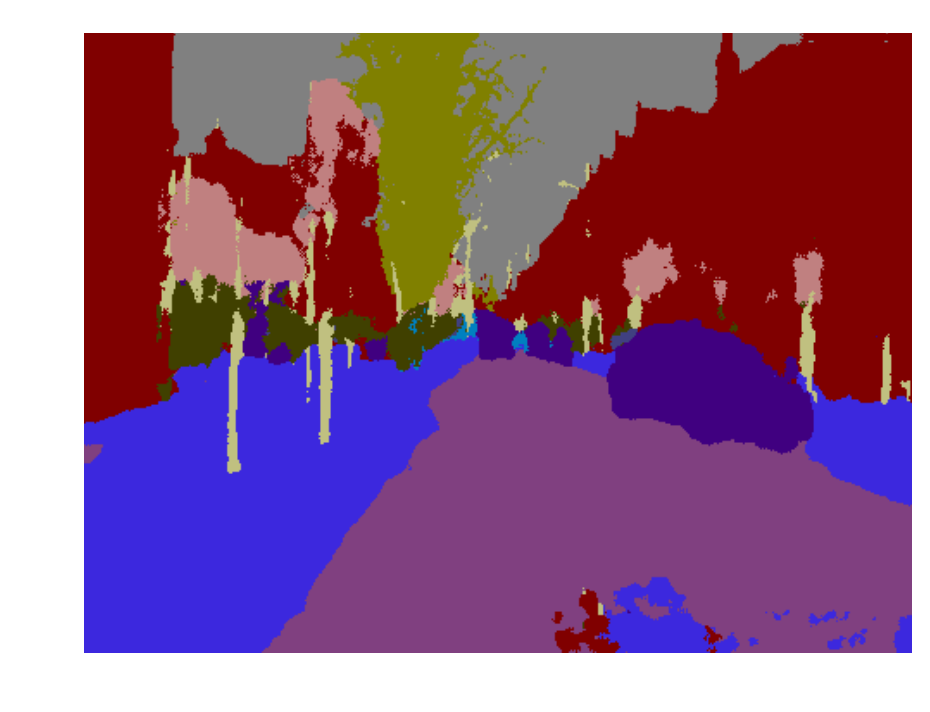}
			\caption{Semantic Segmentation}
		\end{subfigure}
		\begin{subfigure}[t]{0.2\linewidth}
			\includegraphics[width=\linewidth,trim={4cm 3cm 0cm 2cm},clip]{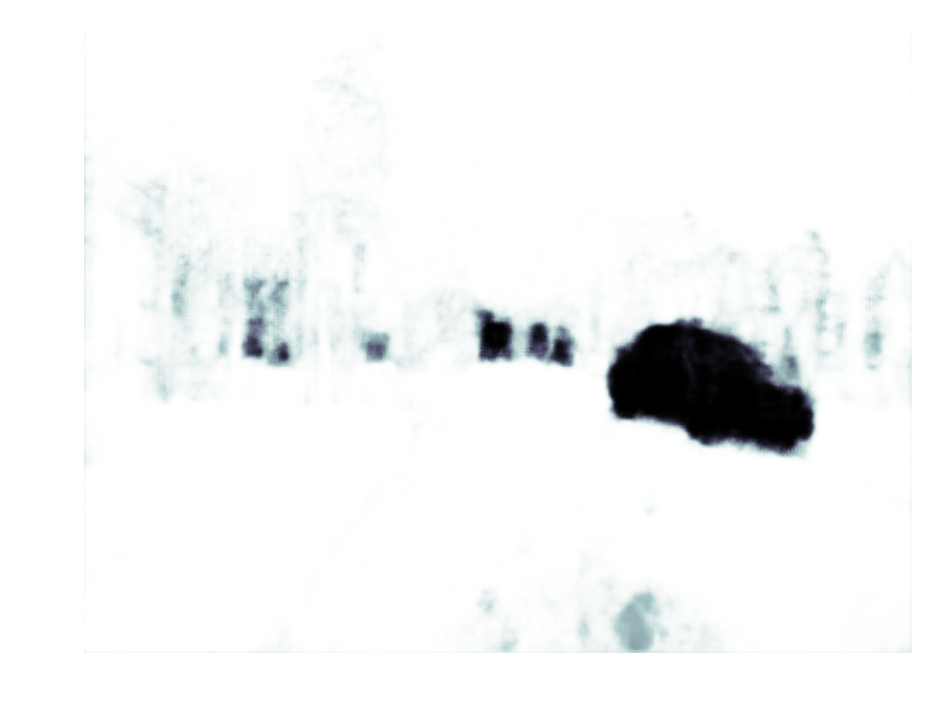}
			\caption{Softmax Uncertainty Car Class}
		\end{subfigure}
		\begin{subfigure}[t]{0.2\linewidth}
			\includegraphics[width=\linewidth,trim={4cm 3cm 0cm 2cm},clip]{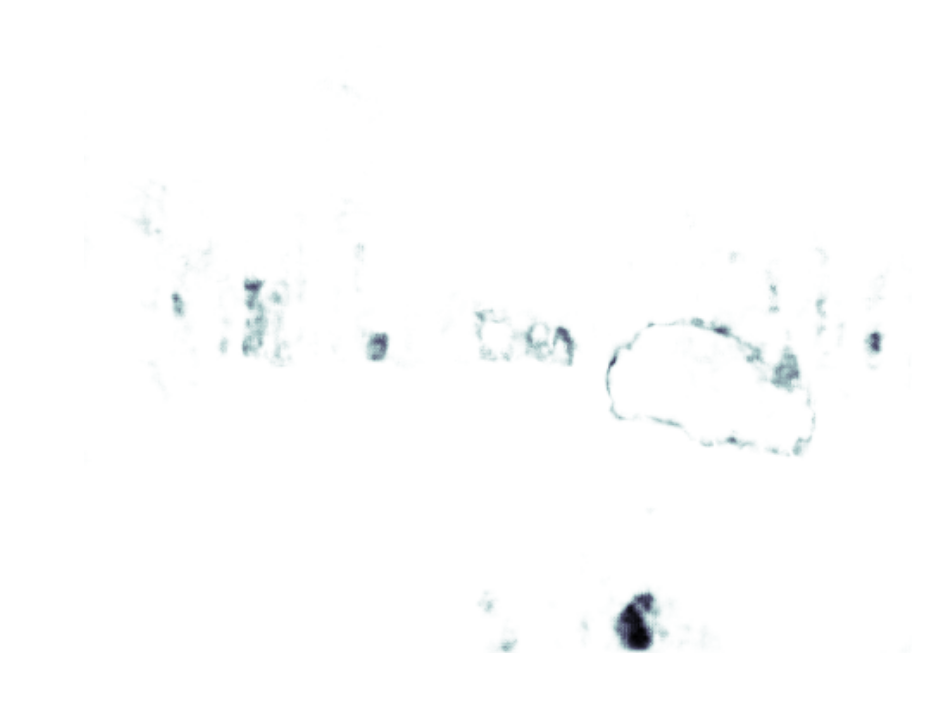}
			\caption{Dropout Uncertainty Car Class}
		\end{subfigure}
		\begin{subfigure}[t]{0.2\linewidth}
			\includegraphics[width=\linewidth,trim={4cm 3cm 0cm 2cm},clip]{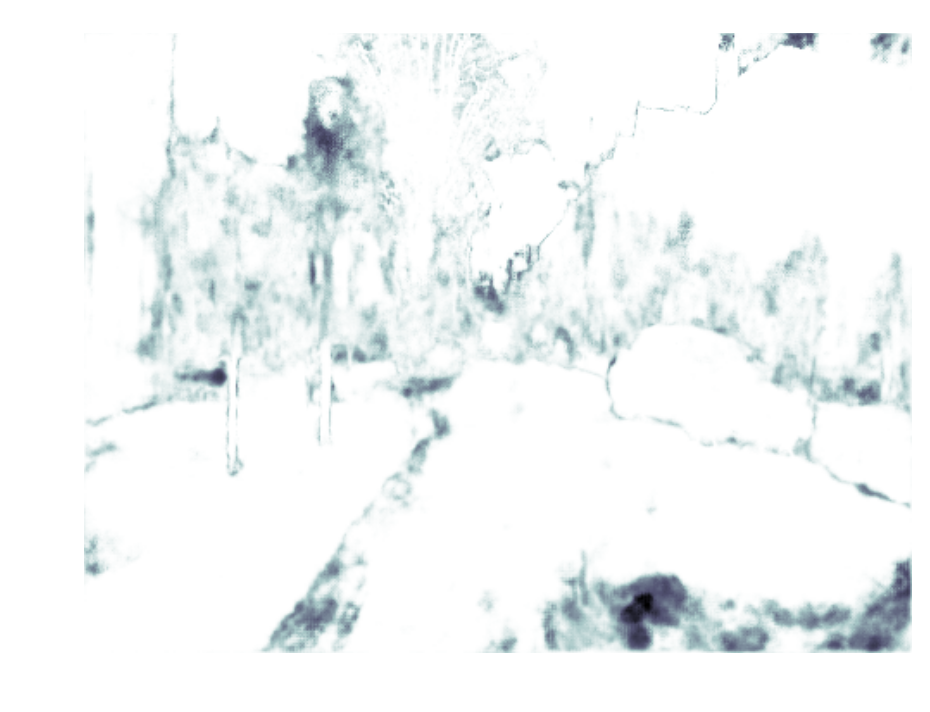}
			\caption{Dropout Uncertainty All Classes}
		\end{subfigure}
	}
	\end{center}
	\caption{\textbf{Comparison of uncertainty with Monte Carlo dropout and uncertainty from softmax regression (c-e: darker colour represents larger value).} This figure shows that softmax regression is only capable of inferring relative probabilities between classes. In contrast, dropout uncertainty can produce an estimate of absolute model uncertainty.}
	\label{fig:uncertainty}
\end{figure*}

We note that the probability distribution from Monte Carlo sampling is significantly different to the \textit{`probabilities'} obtained from a softmax classifier. The softmax function approximates relative probabilities between the class labels, but not an overall measure of the model's uncertainty \cite{Gal2015DropoutB}. Figure \ref{fig:uncertainty} illustrates these differences.

\section{SegNet Architecture}

\begin{figure*}[t]
\begin{center}
\includegraphics[width=\linewidth]{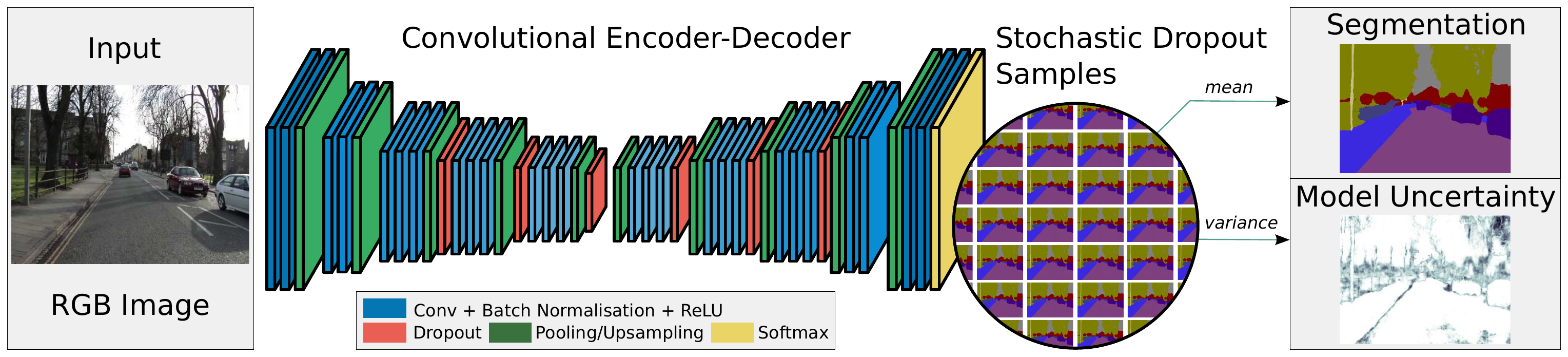}
\end{center}
   \caption{\textbf{A schematic of the Bayesian SegNet architecture.} This diagram shows the entire pipeline for the system which is trained end-to-end in one step with stochastic gradient descent. The encoders are based on the 13 convolutional layers of the VGG-16 network \cite{simonyan2014very}, with the decoder placing them in reverse. The probabilistic output is obtained from Monte Carlo samples of the model with dropout at test time. We take the variance of these softmax samples as the model uncertainty for each class.}
\label{fig:arch}
\end{figure*}

We briefly review the SegNet architecture \cite{badrinarayanan2015segnet} which we modify to produce Bayesian SegNet. SegNet is a deep convolutional encoder decoder architecture which consists of a sequence of non-linear processing layers (encoders) and a corresponding set of decoders followed by a pixel-wise classifier. Typically, each encoder consists of one or more convolutional layers with batch normalisation and a ReLU non-linearity, followed by non-overlapping max-pooling and sub-sampling. The sparse encoding due to the pooling process is upsampled in the decoder using the max-pooling indices in the encoding sequence. This has the important advantage of retaining class boundary details in the segmented images and also reducing the total number of model parameters. The model is trained end to end using stochastic gradient descent.

We take both SegNet \cite{badrinarayanan2015segnet} and a smaller variant termed SegNet-Basic \cite{badrinarayanan2015segnetlayerwise} as our base models. SegNet's encoder is based on the 13 convolutional layers of the VGG-16 network \cite{simonyan2014very} followed by 13 corresponding decoders. SegNet-Basic is a much smaller network with only four layers each for the encoder and decoder with a constant feature size of 64. We use SegNet-Basic as a smaller model for our analysis since it conceptually mimics the larger architecture.

\section{Bayesian SegNet}
\label{sec:bayesian_segnet}

The technique we use to form a probabilistic encoder-decoder architecture is dropout \cite{srivastava2014dropout}, which we use as approximate inference in a Bayesian neural network \cite{Gal2015Bayesian}. We can therefore consider using dropout as a way of getting samples from the posterior distribution of models. Gal and Ghahramani \cite{Gal2015Bayesian} link this technique to variational inference in Bayesian convolutional neural networks with Bernoulli distributions over the network's weights. We leverage this method to perform probabilistic inference over our segmentation model, giving rise to Bayesian SegNet.

For Bayesian SegNet we are interested in finding the posterior distribution over the convolutional weights, $\mathbf{W}$, given our observed training data $\mathbf{X}$ and labels $\mathbf{Y}$.
\begin{equation}
p(\mathbf{W}~|~\mathbf{X},\mathbf{Y})
\end{equation}
In general, this posterior distribution is not tractable, therefore we need to approximate the distribution of these weights \cite{denker1991transforming}. Here we use variational inference to approximate it \cite{graves2011practical}. This technique allows us to learn the distribution over the network's weights, $q(\mathbf{W})$, by minimising the Kullback-Leibler (KL) divergence between this approximating distribution and the full posterior;

\begin{equation}
\text{KL}(q(\mathbf{W})~||~p(\mathbf{W}~|~\mathbf{X},\mathbf{Y})).
\end{equation}
Here, the approximating variational distribution $q(\mathbf{W_i})$ for every $K\times K$ dimensional convolutional layer $i$, with units $j$, is defined as:
\begin{equation}
\begin{split}
\mathbf{b}_{i,j} \sim \text{Bernoulli}(p_i) \text{ for } j = 1, ..., K_{i}, \\
\mathbf{W_i} = \mathbf{M}_i \text{diag}(b_i),
\end{split}
\end{equation}
with $b_i$ vectors of Bernoulli distributed random variables and variational parameters $\mathbf{M_i}$ we obtain the approximate model of the Gaussian process in \cite{Gal2015Bayesian}. The dropout probabilities, $p_i$, could be optimised. However we fix them to the standard probability of dropping a connection as 50\%, i.e. $p_i=0.5$ \cite{srivastava2014dropout}.

In \cite{Gal2015Bayesian} it was shown that minimising the cross entropy loss objective function has the effect of minimising the Kullback-Leibler divergence term. Therefore training the network with stochastic gradient descent will encourage the model to learn a distribution of weights which explains the data well while preventing over-fitting.

We train the model with dropout and sample the posterior distribution over the weights at test time using dropout to obtain the posterior distribution of softmax class probabilities. We take the \textit{mean of these samples for our segmentation prediction} and use the \textit{variance to output model uncertainty for each class}. We take the mean of the per class variance measurements as an overall measure of model uncertainty. We also explored using the \textit{variation ratio} as a measure of uncertainty (i.e. the percentage of samples which agree with the class prediction) however we found this to qualitatively produce a more binary measure of model uncertainty. Fig. \ref{fig:arch} shows a schematic of the segmentation prediction and model uncertainty estimate process.

\subsection{Probabilistic Variants}

\begin{table}[t]
\centering
\tabcolsep=3pt
\label{my-label}
 \resizebox{\linewidth}{!}{
\begin{tabular}{l|ccc|ccc|ccc}
 & \multicolumn{3}{c|}{Weight} & \multicolumn{3}{c|}{Monte Carlo} & \multicolumn{3}{c}{Training} \\
 & \multicolumn{3}{c|}{Averaging} & \multicolumn{3}{c|}{Sampling} & \multicolumn{3}{c}{Fit} \\
Probabilistic Variants & G & C & I/U & G & C & I/U & G & C & I/U\\
\hline \hline
No Dropout				& 82.9 & 62.4 & 46.4 & n/a & n/a & n/a & 94.7 & \textbf{96.2} & \textbf{92.7} \\
Dropout Encoder			& 80.6 & 68.9 & 53.4 & 81.6 & 69.4 & 54.0 & 90.6 & 92.5 & 86.3 \\
Dropout Decoder			& 82.4 & 64.5 & 48.8 & 82.6 & 62.4 & 46.1 & 94.6 & 96.0 & 92.4 \\
Dropout Enc-Dec	& 79.9 & 69.0 & 54.2 & 79.8 & 68.8 & 54.0 & 88.9 & 89.0 & 80.6\\
Dropout Central Enc-Dec	& 81.1 & \textbf{70.6} & \textbf{55.7} & 81.6 & \textbf{70.6} & \textbf{55.8} & 90.4 & 92.3 & 85.9 \\
Dropout Center			& 82.9 & 68.9 & 53.1 & 82.7 & 68.9 & 53.2 & 93.3 & 95.4 & 91.2 \\
Dropout Classifier		& \textbf{84.2} & 62.6 & 46.9 & \textbf{84.2} & 62.6 & 46.8 & \textbf{94.9} & 96.0 & 92.3 \\
\hline
\end{tabular}
}
\caption{\textbf{Architecture Variants for SegNet-Basic on the CamVid dataset} \cite{brostow2009semantic}. We compare the performance of weight averaging against 50 Monte Carlo samples. We quantify performance with three metrics; global accuracy (G), class average accuracy (C) and intersection over union (I/U). Results are shown as percentages (\%). We observe that dropping out every encoder and decoder is too strong a regulariser and results in a lower training fit. The optimal result across all classes is when only the central encoder and decoders are dropped out.}
\label{tbl:bayesvariants}
\end{table}

A fully Bayesian network should be trained with dropout after every convolutional layer. However we found in practice that this was too strong a regulariser, causing the network to learn very slowly. We therefore explored a number of variants that have different configurations of Bayesian or deterministic encoder and decoder units. We note that an encoder unit contains one or more convolutional layers followed by a max pooling layer. A decoder unit contains one or more convolutional layers followed by an upsampling layer. The variants are as follows:

\begin{itemize}[noitemsep]
\item \textbf{Bayesian Encoder}. In this variant we insert dropout after each encoder unit.
\item \textbf{Bayesian Decoder}. In this variant we insert dropout after each decoder unit.
\item \textbf{Bayesian Encoder-Decoder}. In this variant we insert dropout after each encoder and decoder unit.
\item \textbf{Bayesian Center}. In this variant we insert dropout after the deepest encoder, between the encoder and decoder stage.
\item \textbf{Bayesian Central Four Encoder-Decoder}. In this variant we insert dropout after the central four encoder and decoder units.
\item \textbf{Bayesian Classifier}. In this variant we insert dropout after the last decoder unit, before the classifier.
\end{itemize}

For analysis we use the smaller eight layer SegNet-Basic architecture \cite{badrinarayanan2015segnet} and test these Bayesian variants on the CamVid dataset \cite{brostow2009semantic}. We observe qualitatively that all four variants produce similar looking model uncertainty output. That is, they are uncertain near the border of segmentations and with visually ambiguous objects, such as cyclist and pedestrian classes. However, Table \ref{tbl:bayesvariants} shows a difference in quantitative segmentation performance. 

We observe using dropout after all the encoder and decoder units results in a lower training fit and poorer test performance as it is too strong a regulariser on the model. We find that dropping out half of the encoder or decoder units is the optimal configuration. The best configuration is dropping out the deepest half of the encoder and decoder units. We therefore benchmark our Bayesian SegNet results on the Central Enc-Dec variant. For the full 26 layer Bayesian SegNet, we add dropout to the central six encoders and decoders. This is illustrated in Fig. \ref{fig:arch}.

In the lower layers of convolutional networks basic features are extracted, such as edges and corners \cite{zeiler2014visualizing}. These results show that applying Bayesian weights to these layers does not result in a better performance. We believe this is because these low level features are consistent across the distribution of models because they are better modelled with deterministic weights. However, the higher level features that are formed in the deeper layers, such as shape and contextual relationships, are more effectively modelled with Bayesian weights.

\begin{figure}[t]
\begin{center}
	\begin{subfigure}[b]{0.48\linewidth}
        \includegraphics[width=\linewidth]{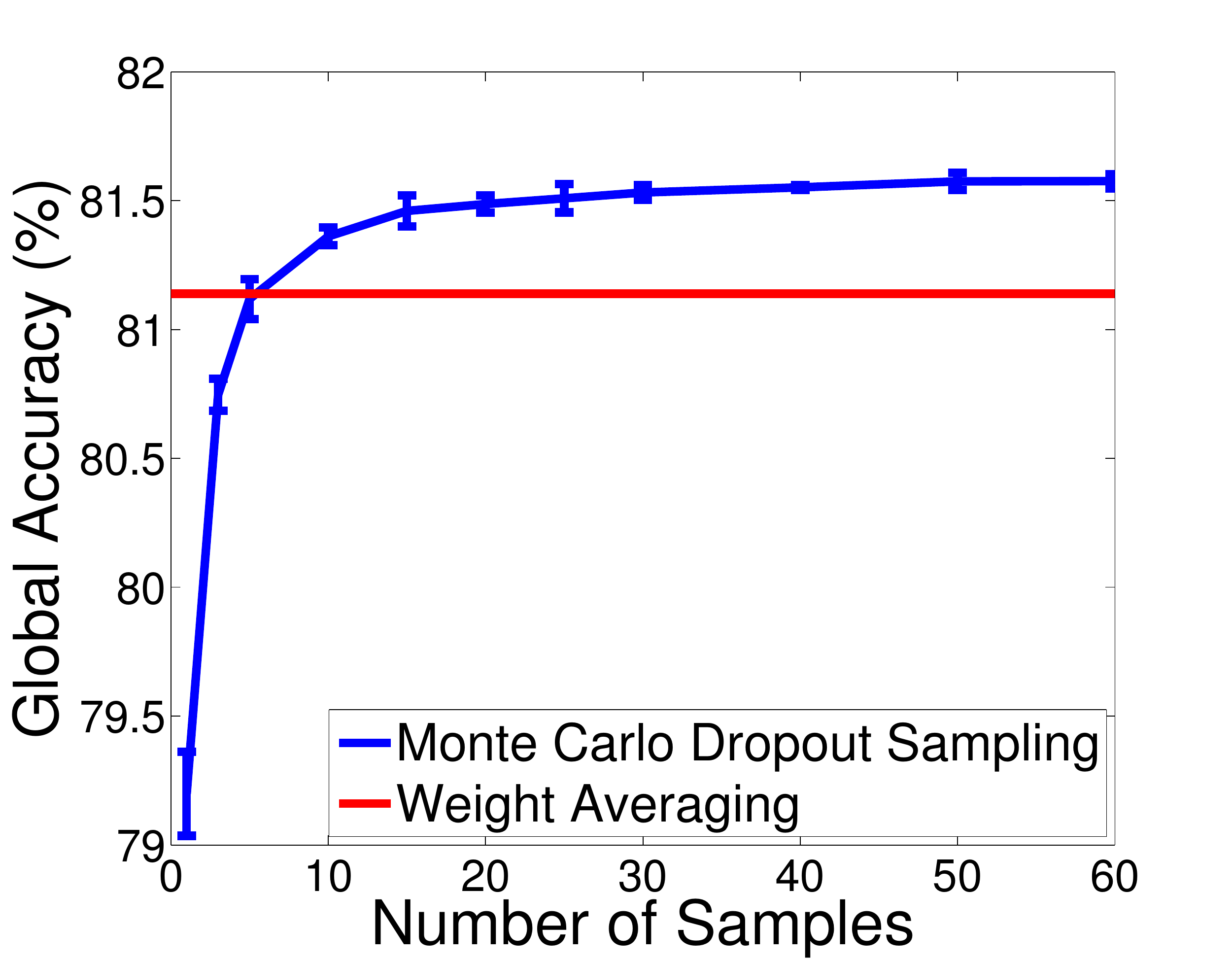}
        \caption{SegNet Basic}
    \end{subfigure}
    	\begin{subfigure}[b]{0.48\linewidth}
        \includegraphics[width=\linewidth]{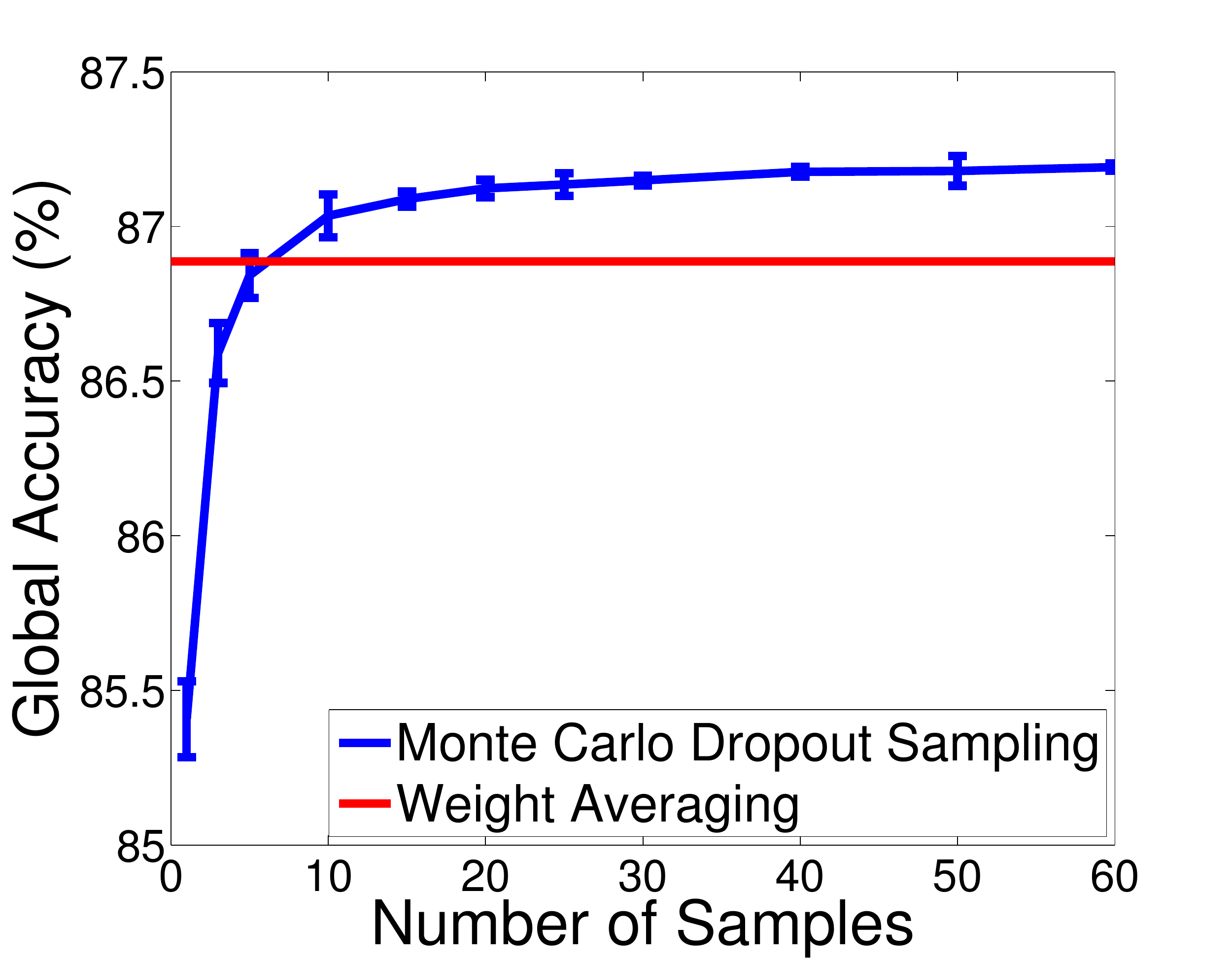}
        \caption{SegNet}
    \end{subfigure}
\end{center}
   \caption{\textbf{Global segmentation accuracy against number of Monte Carlo samples for both SegNet and SegNet-Basic.} Results averaged over 5 trials, with two standard deviation error bars, are shown for the CamVid dataset. This shows that Monte Carlo sampling outperforms the weight averaging technique after approximately 6 samples. Monte Carlo sampling converges after around 40 samples with no further significant improvement beyond this point.}
\label{fig:samples}
\end{figure}

\subsection{Comparing Weight Averaging and Monte Carlo Dropout Sampling}

\begin{table*}[t]
	\resizebox{\textwidth}{!}{
		\small{
			\begin{tabular}{c|c|c|c|c|c|c|c|c|c|c|c|ccc}
				
				\multicolumn{1}{c}{Method}                   & \multicolumn{1}{c}{\rotatebox{90}{Building}} & \multicolumn{1}{c}{\rotatebox{90}{Tree}} & \multicolumn{1}{c}{\rotatebox{90}{Sky}}  & \multicolumn{1}{c}{\rotatebox{90}{Car}}  & \multicolumn{1}{c}{\rotatebox{90}{Sign-Symbol}} & \multicolumn{1}{c}{\rotatebox{90}{Road}} & \multicolumn{1}{c}{\rotatebox{90}{Pedestrian}} & \multicolumn{1}{c}{\rotatebox{90}{Fence}} & \multicolumn{1}{c}{\rotatebox{90}{Column-Pole}} & \multicolumn{1}{c}{\rotatebox{90}{Side-walk}} & \multicolumn{1}{c}{\rotatebox{90}{Bicyclist}} & \multicolumn{1}{c}{\rotatebox{90}{Class avg.}} & \multicolumn{1}{c}{\rotatebox{90}{Global avg.}} & \multicolumn{1}{c}{\rotatebox{90}{Mean I/U}}\\ \hline \hline
				
				SfM+Appearance  \cite{brostow2008segmentation}           & 46.2     & 61.9 & 89.7 & 68.6 & 42.9        & 89.5 & 53.6       & 46.6  & 0.7         & 60.5     & 22.5      & 53.0       & 69.1  & n/a      \\ \hline
				
				Boosting    \cite{sturgess2009combining}              & 61.9     & 67.3 & 91.1 & 71.1 & 58.5        & 92.9 & 49.5       & 37.6  & 25.8        & 77.8     & 24.7      & 59.8       & 76.4  & n/a      \\ \hline
				
				Structured Random Forests \cite{kontschieder2011structured}& \multicolumn{11}{c|}{n/a}                                                                          & 51.4       & 72.5    & n/a    \\ \hline
				
				Neural Decision Forests \cite{rota2014neural}  & \multicolumn{11}{c|}{n/a}                                                                          & 56.1       & 82.1   & n/a     \\ \hline
				
				Local Label Descriptors  \cite{yang2012local}  & 80.7     & 61.5 & 88.8 & 16.4 & n/a         & 98.0 & 1.09       & 0.05  & 4.13        & 12.4     & 0.07      & 36.3       & 73.6    & n/a    \\ \hline
				
				Super Parsing   \cite{tighe2013superparsing}           & 87.0     & 67.1 & 96.9 & 62.7 & 30.1        & 95.9 & 14.7       & 17.9  & 1.7         & 70.0     & 19.4      & 51.2       & 83.3    & n/a    \\ \hline
				
				Boosting+Detectors+CRF \cite{ladicky2010and}   & 81.5     & 76.6 & 96.2 & 78.7 & 40.2        & 93.9 & 43.0       & 47.6  & 14.3        & 81.5     & 33.9      & 62.5       & 83.8    & n/a    \\ \hline
				
				SegNet-Basic (layer-wise training \cite{badrinarayanan2015segnetlayerwise})         & 75.0     & 84.6 & 91.2 & 82.7 & 36.9        & 93.3 & 55.0       & 37.5  & 44.8        & 74.1     & 16.0      & 62.9       & 84.3   & n/a     \\ \hline
				SegNet-Basic \cite{badrinarayanan2015segnet} &  80.6    & 72.0  & 93.0 & 78.5 & 21.0   & 94.0 &   62.5     & 31.4  &     36.6    &  74.0   & 42.5      &   62.3    & 82.8   &   46.3   \\ \hline
				SegNet \cite{badrinarayanan2015segnet}  &  \textbf{88.0 }   & \textbf{87.3} & 92.3  & 80.0 & 29.5   & 97.6 &  57.2      & 49.4  &  27.8       &  84.8   &  30.7     &   65.9    &  88.6  & 50.2     \\ \hline
				FCN 8 \cite{long2014fully} & \multicolumn{11}{c|}{n/a} & 64.2 & 83.1 & 52.0 \\ \hline
				DeconvNet \cite{noh2015learning} & \multicolumn{11}{c|}{n/a} & 62.1 & 85.9 & 48.9 \\ \hline
				DeepLab-LargeFOV-DenseCRF \cite{chen2014semantic} & \multicolumn{11}{c|}{n/a} & 60.7 & \textbf{89.7} & 54.7 \\ \hline
				
				\multicolumn{15}{c}{\emph{Bayesian SegNet Models in this work:}} \\ \hline

				Bayesian SegNet-Basic 	& 75.1 & 68.8 & 91.4 & 77.7 & 52.0 & 92.5 & 71.5 & 44.9 & \textbf{52.9} & 79.1 & 69.6 & 70.5 & 81.6 & 55.8 \\ \hline
				Bayesian SegNet 			& 80.4 & 85.5 & 90.1 & \textbf{86.4} & \textbf{67.9} & 93.8 & \textbf{73.8} & \textbf{64.5} & 50.8 & \textbf{91.7} & \textbf{54.6} & \textbf{76.3} & 86.9 & \textbf{63.1} \\
				
			\end{tabular}
		}}
		\vspace*{0.1cm}
		\caption{\textbf{Quantitative results on CamVid} \cite{brostow2009semantic} consisting of 11 road scene categories. Bayesian SegNet outperforms all other methods, including shallow methods which utilise depth, video and/or CRF's, and more contemporary deep methods. Particularly noteworthy are the significant improvements in accuracy for the smaller/thinner classes.
		}
		\label{tbl:camvid}
	\end{table*}

Monte Carlo dropout sampling qualitatively allows us to understand the model uncertainty of the result. However, for segmentation, we also want to understand the quantitative difference between sampling with dropout and using the weight averaging technique proposed by \cite{srivastava2014dropout}. Weight averaging proposes to remove dropout at test time and scale the weights proportionally to the dropout percentage. Fig. \ref{fig:samples} shows that Monte Carlo sampling with dropout performs better than weight averaging after approximately 6 samples. We also observe no additional performance improvement beyond approximately 40 samples. Therefore the weight averaging technique produces poorer segmentation results, in terms of global accuracy, in addition to being unable to provide a measure of model uncertainty. However, sampling comes at the expense of inference time, but when computed in parallel on a GPU this cost can be reduced for practical applications.

\subsection{Training and Inference}

Following \cite{badrinarayanan2015segnet} we train SegNet with median frequency class balancing using the formula proposed by Eigen and Fergus \cite{eigen2014predicting}. We use batch normalisation layers after every convolutional layer \cite{ioffe2015batch}. We compute batch normalisation statistics across the training dataset and use these at test time. We experimented with computing these statistics while using dropout sampling. However we experimentally found that computing them with weight averaging produced better results.

We implement Bayesian SegNet using the Caffe library \cite{jia2014caffe} and release the source code and trained models for public evaluation \footnote{An online demo and source code can be found on our project webpage \href{mi.eng.cam.ac.uk/projects/segnet/}{mi.eng.cam.ac.uk/projects/segnet/}}. We train the whole system end-to-end using stochastic gradient descent with a base learning rate of 0.001 and weight decay parameter equal to 0.0005. We train the network until convergence when we observe no further reduction in training loss.

\section{Experiments}
\label{sec:results}

We quantify the performance of Bayesian SegNet on three different benchmarks using our Caffe implementation. Through this process we demonstrate the efficacy of Bayesian SegNet for a wide variety of scene segmentation tasks which have practical applications. CamVid \cite{brostow2009semantic} is a road scene understanding dataset which has applications for autonomous driving. SUN RGB-D \cite{song2015sun} is a very challenging and large dataset of indoor scenes which is important for domestic robotics. Finally, Pascal VOC 2012 \cite{everingham2010pascal} is a RGB dataset for object segmentation.

\begin{figure*}[p]
\begin{center}
\makebox[\linewidth][c]{
\includegraphics[height=0.125\linewidth]{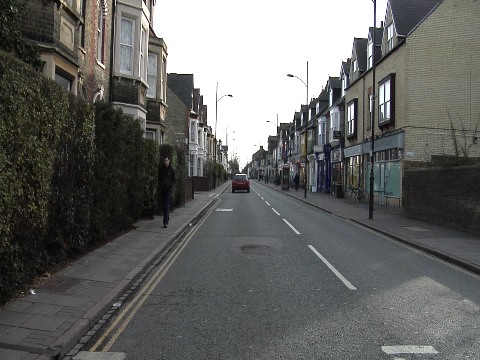}
\includegraphics[height=0.125\linewidth]{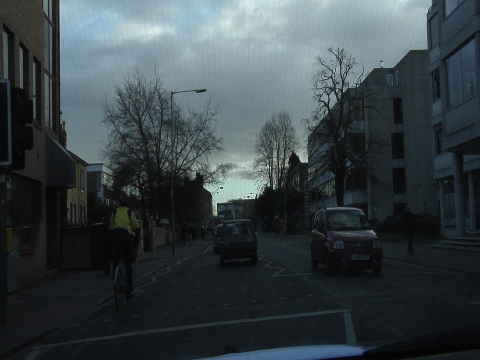}
\includegraphics[height=0.125\linewidth]{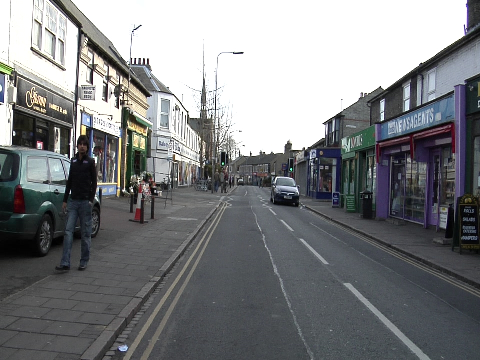}
\includegraphics[height=0.125\linewidth]{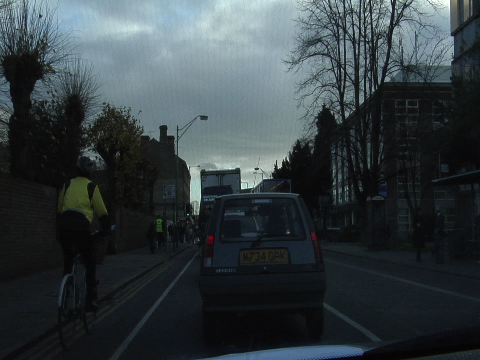}
\includegraphics[height=0.125\linewidth]{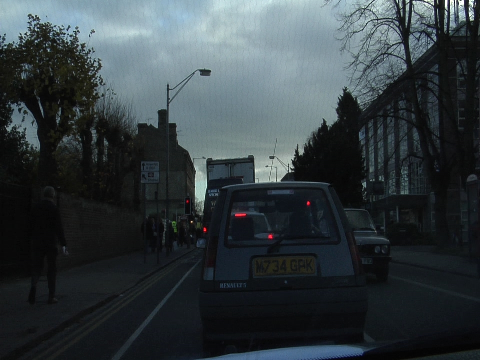}
\includegraphics[height=0.125\linewidth]{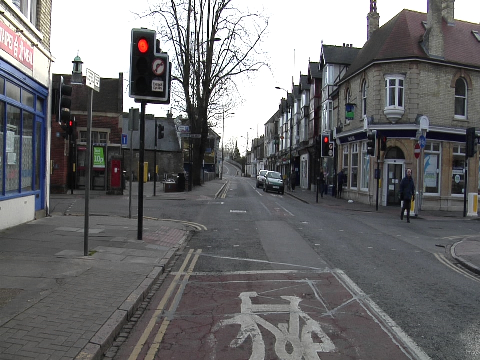}
}
\makebox[\linewidth][c]{
\includegraphics[height=0.125\linewidth]{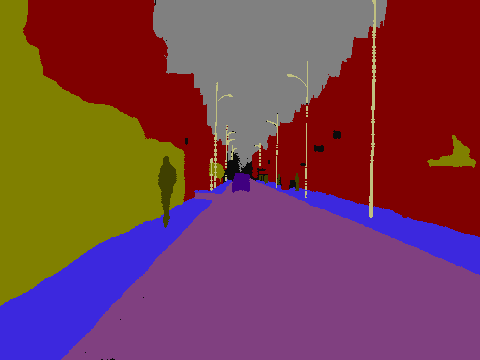}
\includegraphics[height=0.125\linewidth]{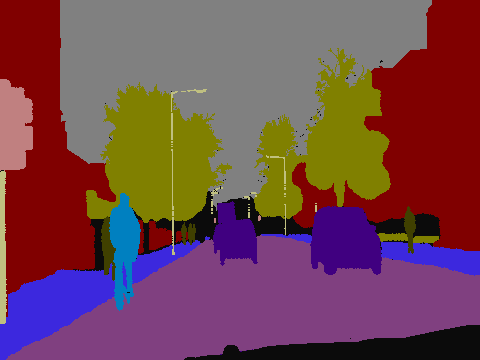}
\includegraphics[height=0.125\linewidth]{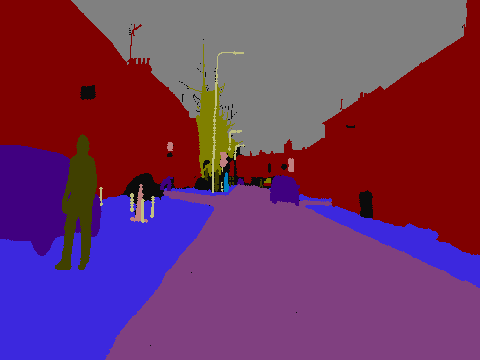}
\includegraphics[height=0.125\linewidth]{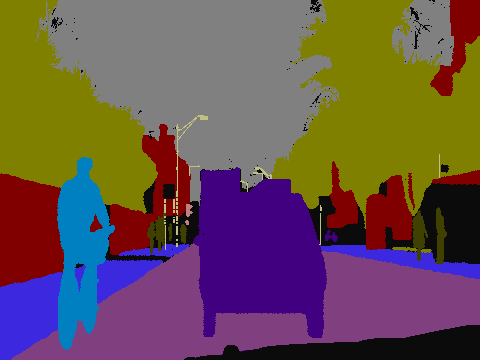}
\includegraphics[height=0.125\linewidth]{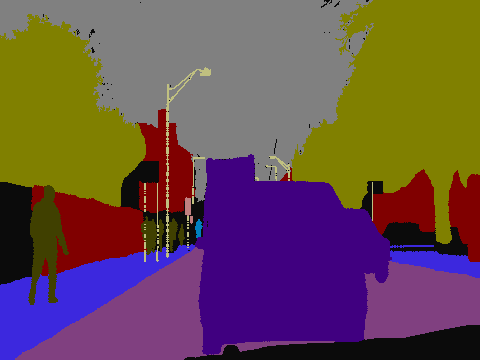}
\includegraphics[height=0.125\linewidth]{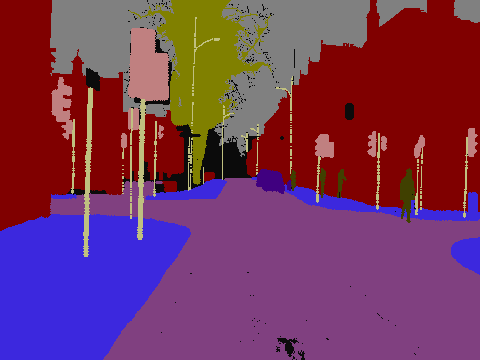}
}
\makebox[\linewidth][c]{
\includegraphics[height=0.125\linewidth]{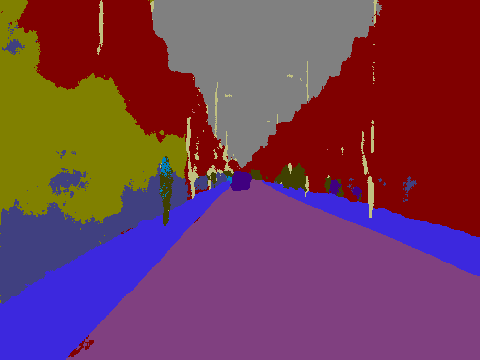}
\includegraphics[height=0.125\linewidth]{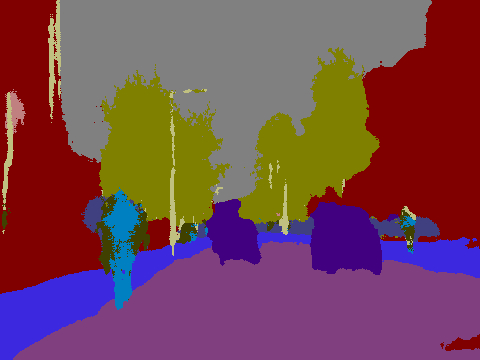}
\includegraphics[height=0.125\linewidth]{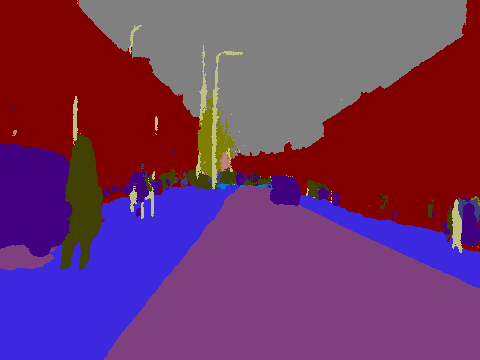}
\includegraphics[height=0.125\linewidth]{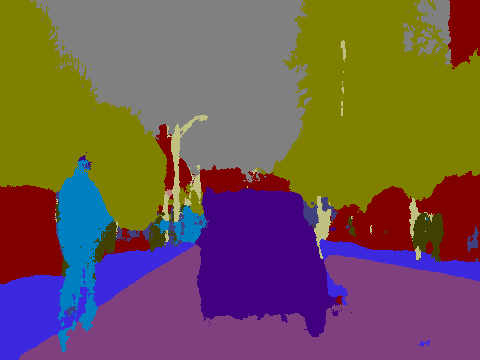}
\includegraphics[height=0.125\linewidth]{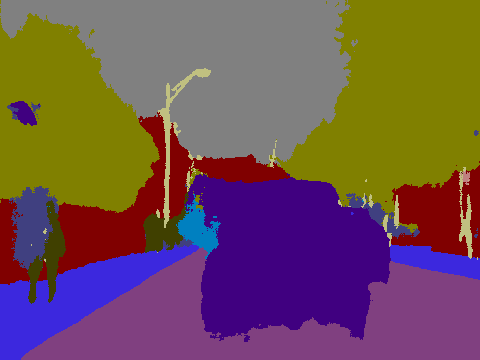}
\includegraphics[height=0.125\linewidth]{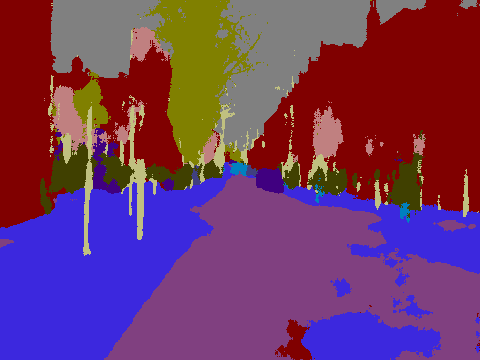}
}
\makebox[\linewidth][c]{
\includegraphics[height=0.125\linewidth]{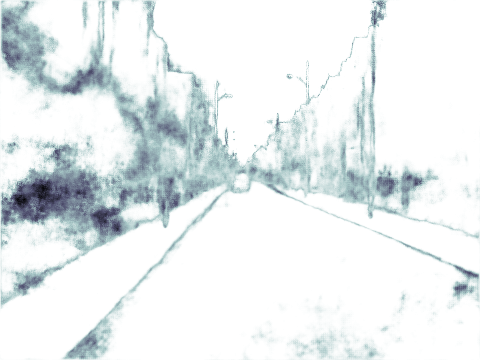}
\includegraphics[height=0.125\linewidth]{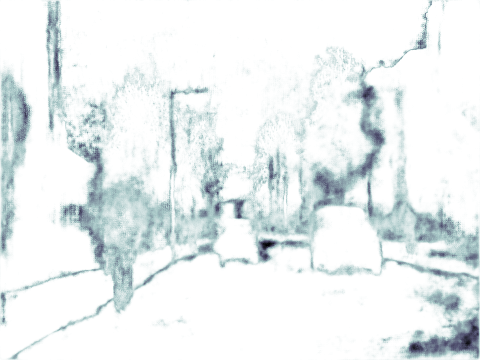}
\includegraphics[height=0.125\linewidth]{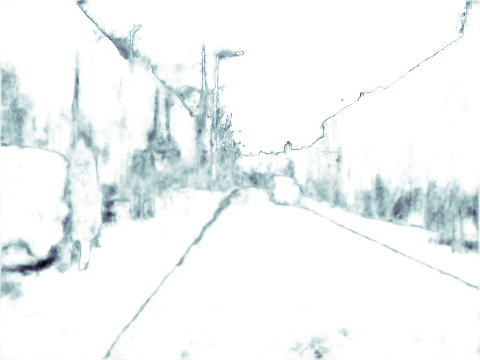}
\includegraphics[height=0.125\linewidth]{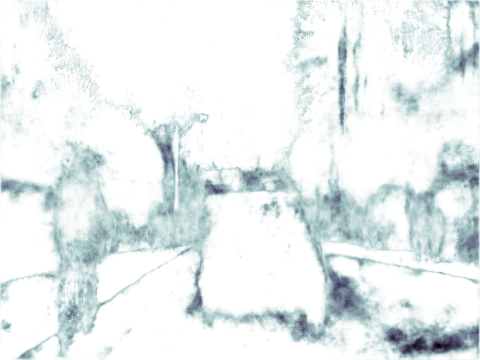}
\includegraphics[height=0.125\linewidth]{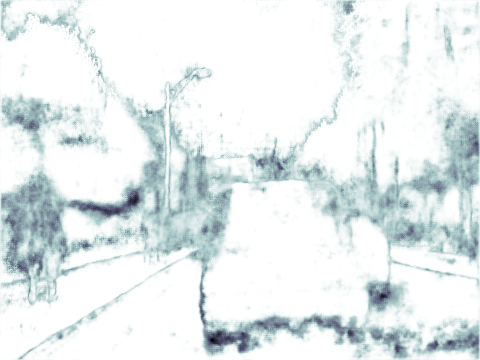}
\includegraphics[height=0.125\linewidth]{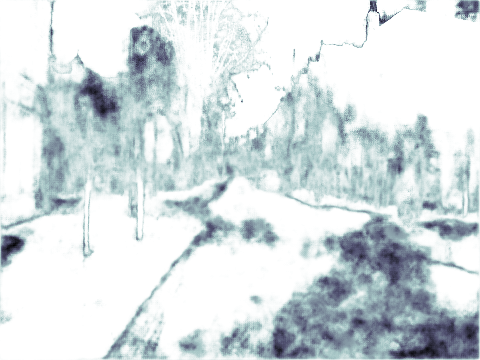}
}
\end{center}
   \caption{\textbf{Bayesian SegNet results on CamVid road scene understanding dataset \cite{brostow2009semantic}.} The top row is the input image, with the ground truth shown in the second row. The third row shows Bayesian SegNet's segmentation prediction, with overall model uncertainty, averaged across all classes, in the bottom row (with darker colours indicating more uncertain predictions). In general, we observe high quality segmentation, especially on more difficult classes such as poles, people and cyclists. Where SegNet produces an incorrect class label we often observe a high model uncertainty.}
\label{fig:qual_camvid}

\begin{center}
\makebox[\linewidth][c]{
\includegraphics[height=0.125\linewidth]{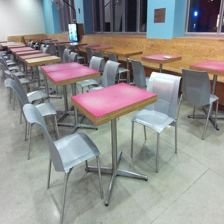}
\includegraphics[height=0.125\linewidth]{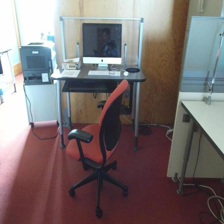}
\includegraphics[height=0.125\linewidth]{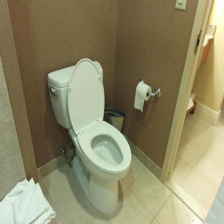}
\includegraphics[height=0.125\linewidth]{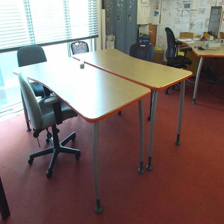}
\includegraphics[height=0.125\linewidth]{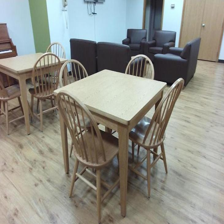}
\includegraphics[height=0.125\linewidth]{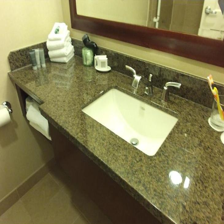}
\includegraphics[height=0.125\linewidth]{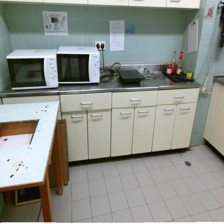}
\includegraphics[height=0.125\linewidth]{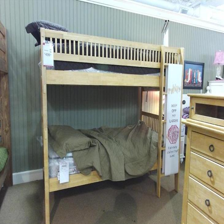}
}
\makebox[\linewidth][c]{
\includegraphics[height=0.125\linewidth]{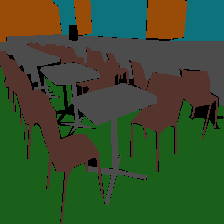}
\includegraphics[height=0.125\linewidth]{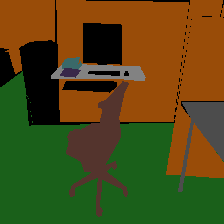}
\includegraphics[height=0.125\linewidth]{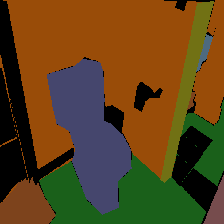}
\includegraphics[height=0.125\linewidth]{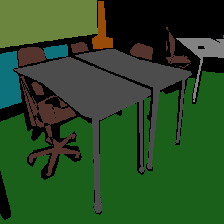}
\includegraphics[height=0.125\linewidth]{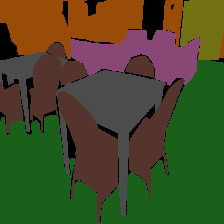}
\includegraphics[height=0.125\linewidth]{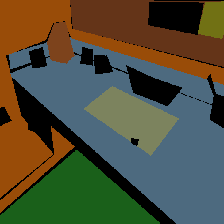}
\includegraphics[height=0.125\linewidth]{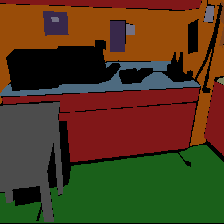}
\includegraphics[height=0.125\linewidth]{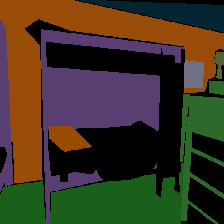}
}
\makebox[\linewidth][c]{
\includegraphics[height=0.125\linewidth]{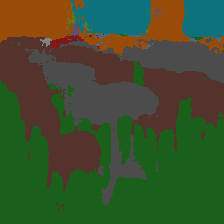}
\includegraphics[height=0.125\linewidth]{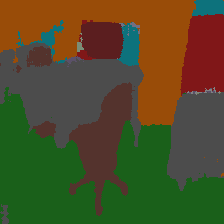}
\includegraphics[height=0.125\linewidth]{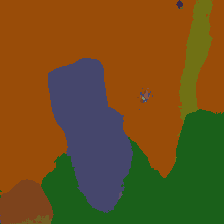}
\includegraphics[height=0.125\linewidth]{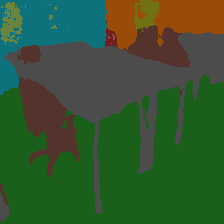}
\includegraphics[height=0.125\linewidth]{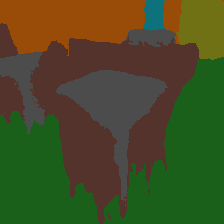}
\includegraphics[height=0.125\linewidth]{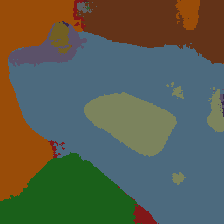}
\includegraphics[height=0.125\linewidth]{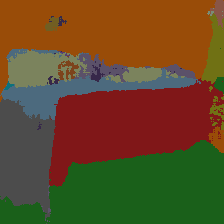}
\includegraphics[height=0.125\linewidth]{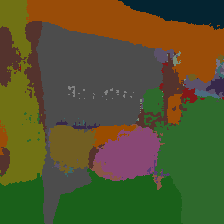}
}
\makebox[\linewidth][c]{
\includegraphics[height=0.125\linewidth]{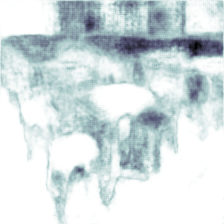}
\includegraphics[height=0.125\linewidth]{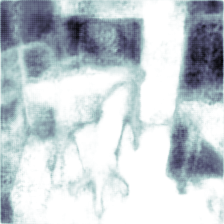}
\includegraphics[height=0.125\linewidth]{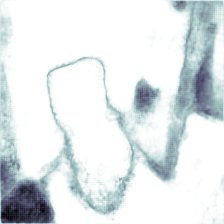}
\includegraphics[height=0.125\linewidth]{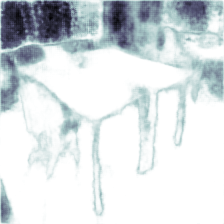}
\includegraphics[height=0.125\linewidth]{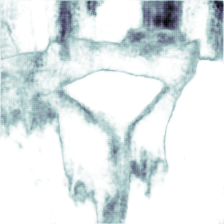}
\includegraphics[height=0.125\linewidth]{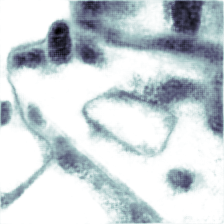}
\includegraphics[height=0.125\linewidth]{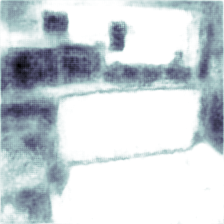}
\includegraphics[height=0.125\linewidth]{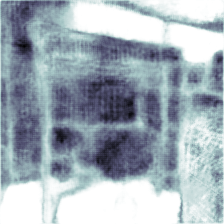}
}
\end{center}
   \caption{\textbf{Bayesian SegNet results on the SUN RGB-D indoor scene understanding dataset \cite{song2015sun}.} The top row is the input image, with the ground truth shown in the second row. The third row shows Bayesian SegNet's segmentation prediction, with overall model uncertainty, averaged across all classes, in the bottom row (with darker colours indicating more uncertain predictions). Bayesian SegNet uses only RGB input and is able to accurately segment 37 classes in this challenging dataset. Note that often parts of an image do not have ground truth labels and these are shown in black colour.}
\label{fig:qual_sun}
\end{figure*}

\subsection{CamVid}

CamVid is a road scene understanding dataset with 367 training images and 233 testing images of day and dusk scenes \cite{brostow2009semantic}. The challenge is to segment $11$ classes such as road, building, cars, pedestrians, signs, poles, side-walk etc. We resize images to 360x480 pixels for training and testing of our system.

Table \ref{tbl:camvid} shows our results and compares them to previous benchmarks. We compare to methods which utilise depth and motion cues. Additionally we compare to other prominent deep learning architectures. Bayesian SegNet obtains the highest overall class average and mean intersection over union score by a significant margin. We set a new benchmark on 7 out of the 11 classes. Qualitative results can be viewed in Fig. \ref{fig:qual_camvid}.

\subsection{Scene Understanding (SUN)}

\begin{table}[t]
	\centering
	\begin{tabular}{c|ccc}
		{Method} & {G} & {C} & {I/U} \\ \hline \hline
		\multicolumn{4}{c}{RGB} \\ \hline
		Liu \emph{et~al.}  \cite{liu2008sift} & n/a & 9.3 & n/a \\ \hline
		FCN 8 \cite{long2014fully} & 68.2 & 38.4 & 27.4 \\ \hline
		DeconvNet \cite{noh2015learning} & 66.1 & 32.3 & 22.6 \\ \hline
		DeepLab-LargeFOV-DenseCRF \cite{chen2014semantic} & 67.0 & 33.0 & 24.1 \\ \hline
		SegNet \cite{badrinarayanan2015segnet} & 70.3 & 35.6 & 22.1 \\ \hline
		Bayesian SegNet \emph{(this work)} & \textbf{71.2} & \textbf{45.9} & \textbf{30.7} \\ \hline
		\multicolumn{4}{c}{RGB-D} \\ \hline
		Liu \emph{et~al.}   \cite{liu2008sift} & n/a & 10.0 & n/a \\ \hline
		Ren et. al \cite{ren2012rgb} & n/a & 36.3 & n/a \\ \hline
	\end{tabular}
	\vspace*{0.1cm}
	\caption{\textbf{SUN Indoor Scene Understanding.} Quantitative comparison on the SUN RGB-D dataset \cite{song2015sun} which consists of 5050 test images of indoor scenes with 37 classes. SegNet RGB based predictions have a high global accuracy and out-perform all previous benchmarks, including those which use depth modality.}
	\label{tbl:SUNRGBDtest}
\end{table}

\begin{table}[t]
	\centering
	\begin{tabular}{c|ccc}
		{Method} & {G} & {C} & {I/U} \\ \hline \hline
		\multicolumn{4}{c}{RGB} \\ \hline
		FCN-32s RGB \cite{long2014fully} & 60.0 & 42.2 & 29.2 \\ \hline
		SegNet \cite{badrinarayanan2015segnet} & 66.1 & 36.0 & 23.6 \\ \hline
		Bayesian SegNet \emph{(this work)} & \textbf{68.0} & 45.8 & 32.4 \\ \hline
		\multicolumn{4}{c}{RGB-D} \\ \hline
		Gupta et al. \cite{gupta2014learning} & 60.3 & - & 28.6 \\ \hline
		FCN-32s RGB-D \cite{long2014fully} & 61.5 & 42.4 & 30.5 \\ \hline
		Eigen et al. \cite{eigen2014predicting} & 65.6 & 45.1 & - \\ \hline
		\multicolumn{4}{c}{RGB-HHA} \\ \hline
		FCN-16s RGB-HHA \cite{long2014fully} & 65.4 & \textbf{46.1} & \textbf{34.0} \\ \hline
	\end{tabular}
	\vspace*{0.1cm}
	\caption{\textbf{NYU v2.} Results for the NYUv2 RGB-D dataset \cite{silberman2012indoor} which consists of 654 test images. Bayesian SegNet is the top performing RGB method.}
	\label{tbl:nyutest}
\end{table}

SUN RGB-D \cite{song2015sun} is a very challenging and large dataset of indoor scenes with $5285$ training and $5050$ testing images. The images are captured by different sensors and hence come in various resolutions. The task is to segment $37$ indoor scene classes including wall, floor, ceiling, table, chair, sofa etc. This task is difficult because object classes come in various shapes, sizes and in different poses with frequent partial occlusions. These factors make this one of the hardest segmentation challenges. For our model, we resize the input images for training and testing to 224x224 pixels. Note that we only use RGB input to our system.  Using the depth modality would necessitate architectural modifications and careful post-processing to fill-in missing depth measurements. This is beyond the scope of this paper. 

Table \ref{tbl:SUNRGBDtest} shows our results on this dataset compared to other methods. Bayesian SegNet outperforms all previous benchmarks, including those which use depth modality. We also note that an earlier benchmark dataset, NYUv2 \cite{silberman2012indoor}, is included as part of this dataset, and Table \ref{tbl:nyutest} shows our evaluation on this subset. Qualitative results can be viewed in Fig. \ref{fig:qual_sun}.

\subsection{Pascal VOC}
\label{sec:pascal}

\begin{table}[]
	\centering
	\tabcolsep=2pt
	\resizebox{\linewidth}{!}{
		\begin{tabular}{c|c|c|c}
			& Parameters & \multicolumn{2}{c}{Pascal VOC Test IoU} \\
			Method & (Millions) & Non-Bayesian & Bayesian \\ \hline \hline
			Dilation Network \cite{YuKoltun2016} & 140.8 & 71.3 & 73.1 \\ \hline
			FCN-8 \cite{long2014fully} & 134.5 & 62.2 & 65.4\\ \hline
			SegNet \cite{badrinarayanan2015segnet} & 29.45 & 59.1 & 60.5 \\ \hline
		\end{tabular}
	}
	\vspace*{0.1cm}
	\caption{\textbf{Pascal VOC12 \cite{everingham2010pascal} test results} evaluated from the online evaluation server. We compare to competing deep learning architectures. Bayesian SegNet is considerably smaller but achieves a competitive accuracy to other methods. We also evaluate FCN \cite{long2014fully} and Dilation Network (front end) \cite{YuKoltun2016} with Monte Carlo dropout sampling. We observe an improvement in segmentation performance across all three deep learning models when using the Bayesian approach. This demonstrates this method's applicability in general. Additional results available on the leaderboard \url{host.robots.ox.ac.uk:8080/leaderboard}}
	\label{tbl:pascal}
\end{table}

The Pascal VOC12 segmentation challenge \cite{everingham2010pascal} consists of segmenting a 20 salient object classes from a widely varying background class. For our model, we resize the input images for training and testing to 224x224 pixels. We train on the 12031 training images and 1456 testing images, with scores computed remotely on a test server. Table \ref{tbl:pascal} shows our results compared to other methods, with qualitative results in Fig. \ref{fig:qual_pascal}.

This dataset is unlike the segmentation for scene understanding benchmarks described earlier which require learning both classes and their spatial context. A number of techniques have been proposed based on this challenge which are increasingly more accurate and complex \footnote{See the full leader board at \url{http://host.robots.ox.ac.uk:8080/leaderboard}}. Our efforts in this benchmarking experiment have not been diverted towards attaining the top rank by either using multi-stage training \cite{long2014fully}, other datasets for pre-training such as MS-COCO \cite{lin2014microsoft}, training and inference aids such as object proposals \cite{noh2015learning} or post-processing using CRF based methods \cite{chen2014semantic,zheng2015conditional}. Although these supporting techniques clearly have value towards increasing the performance it unfortunately does not reveal the true performance of the deep architecture which is the \textit{core segmentation engine}. It however does indicate that some of the large deep networks are difficult to train end-to-end on this task even with pre-trained encoder weights. Therefore, to encourage more controlled benchmarking, we trained Bayesian SegNet end-to-end without other aids and report this performance.

\subsection{General Applicability}
\label{sec:generalapplicable}

To demonstrate the general applicability of this method, we also apply it to other deep learning architectures trained with dropout; FCN \cite{long2014fully} and Dilation Network \cite{YuKoltun2016}. We select these state-of-the-art methods as they are already trained by their respective authors using dropout. We take their trained open source models off the shelf, and evaluate them using 50 Monte Carlo dropout samples. Table \ref{tbl:pascal} shows the mean IoU result of these methods evaluated as Bayesian Neural Networks, as computed by the online evaluation server.

This shows the general applicability of our method. By leveraging this underlying Bayesian framework our method obtains 2-3\% improvement across this range of architectures.

\begin{table}[]
	\centering
		\begin{tabular}{c|c|c}
			Percentile & \multicolumn{2}{c}{Pixel-Wise Classification Accuracy} \\
			Confidence & CamVid & SUN RGBD \\ \hline \hline
			90 & 99.7 & 97.6 \\ \hline
			50 & 98.5 & 92.3 \\ \hline
			10 & 89.5 & 79.0 \\ \hline
			0 & 86.7 & 75.4 \\ \hline
		\end{tabular}
	\vspace*{0.1cm}
	\caption{\textbf{Bayesian SegNet’s accuracy as a function of confidence} for the 90th percentile (10\% most confident pixels) through to the 0th percentile (all pixels). This shows uncertainty is an effective measure of prediction accuracy.}
	\label{tbl:confidence_percentiles}
\end{table}

\subsection{Understanding Model Uncertainty}
\label{sec:uncertainty}

\begin{figure*}[t]
	\begin{center}
		\makebox[\linewidth][c]{
			\includegraphics[height=0.125\linewidth]{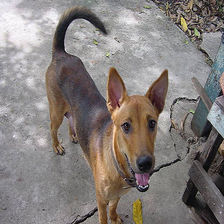}
			\includegraphics[height=0.125\linewidth]{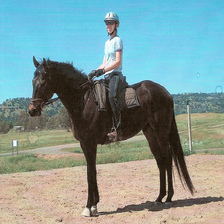}
			\includegraphics[height=0.125\linewidth]{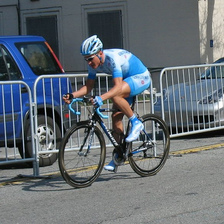}
			\includegraphics[height=0.125\linewidth]{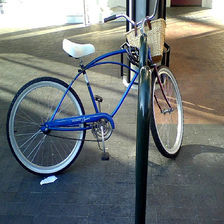}
			\includegraphics[height=0.125\linewidth]{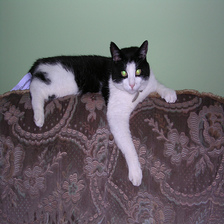}
			\includegraphics[height=0.125\linewidth]{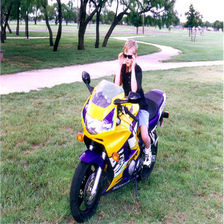}
			\includegraphics[height=0.125\linewidth]{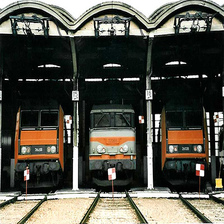}
			\includegraphics[height=0.125\linewidth]{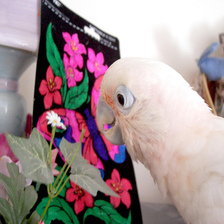}
		}
		\makebox[\linewidth][c]{
			\includegraphics[height=0.125\linewidth]{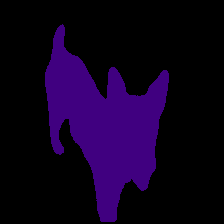}
			\includegraphics[height=0.125\linewidth]{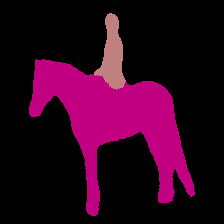}
			\includegraphics[height=0.125\linewidth]{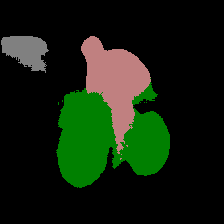}
			\includegraphics[height=0.125\linewidth]{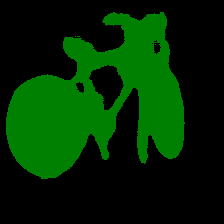}
			\includegraphics[height=0.125\linewidth]{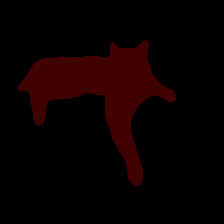}
			\includegraphics[height=0.125\linewidth]{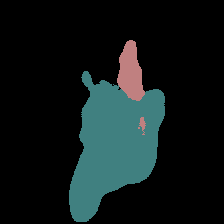}
			\includegraphics[height=0.125\linewidth]{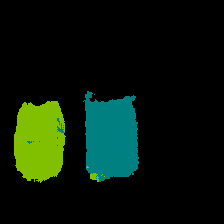}
			\includegraphics[height=0.125\linewidth]{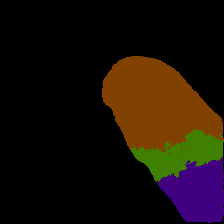}
		}
		\makebox[\linewidth][c]{
			\includegraphics[height=0.125\linewidth]{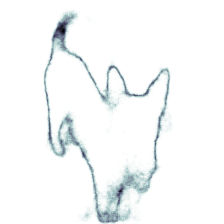}
			\includegraphics[height=0.125\linewidth]{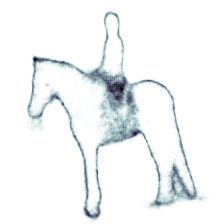}
			\includegraphics[height=0.125\linewidth]{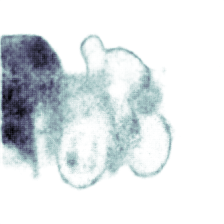}
			\includegraphics[height=0.125\linewidth]{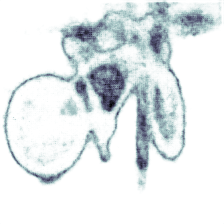}
			\includegraphics[height=0.125\linewidth]{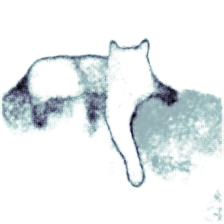}
			\includegraphics[height=0.125\linewidth]{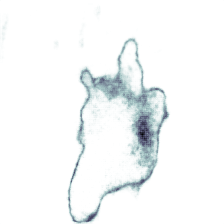}
			\includegraphics[height=0.125\linewidth]{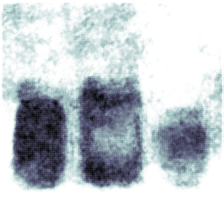}
			\includegraphics[height=0.125\linewidth]{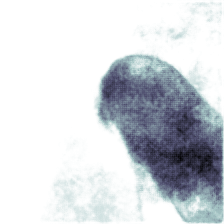}
		}
	\end{center}
	\caption{\textbf{Bayesian SegNet results on the Pascal VOC 2012 dataset \cite{everingham2010pascal}.} The top row is the input image. The middle row shows Bayesian SegNet's segmentation prediction, with overall model uncertainty averaged across all classes in the bottom row (darker colours indicating more uncertain predictions). Ground truth is not publicly available for these test images.}
	\label{fig:qual_pascal}
\end{figure*}

\textbf{Qualitative observations.} Fig. \ref{fig:qual_camvid} shows segmentations and model uncertainty results from Bayesian SegNet on CamVid Road Scenes \cite{brostow2009semantic}. Fig. \ref{fig:qual_sun} shows SUN RGB-D Indoor Scene Understanding \cite{song2015sun} results and Fig. \ref{fig:qual_pascal} has Pascal VOC \cite{everingham2010pascal} results. These figures show the qualitative performance of Bayesian SegNet. We observe that segmentation predictions are smooth, with a sharp segmentation around object boundaries. These results also show that when the model predicts an incorrect label, the model uncertainty is generally very high. More generally, we observe that a high model uncertainty is predominantly caused by three situations.

Firstly, at class boundaries the model often displays a high level of uncertainty. This reflects the ambiguity surrounding the definition of defining where these labels transition. The Pascal results clearly illustrated this in Fig. \ref{fig:qual_pascal}.

Secondly, objects which are visually difficult to identify often appear uncertain to the model. This is often the case when objects are occluded or at a distance from the camera.

The third situation causing model uncertainty is when the object appears visually ambiguous to the model. As an example, cyclists in the CamVid results (Fig. \ref{fig:qual_camvid}) are visually similar to pedestrians, and the model often displays uncertainty around them. We observe similar results with visually similar classes in SUN (Fig. \ref{fig:qual_sun}) such as chair and sofa, or bench and table. In Pascal this is often observed between cat and dog, or train and bus classes.

\begin{figure}[t]
	\begin{center}
		\includegraphics[width=0.7\linewidth]{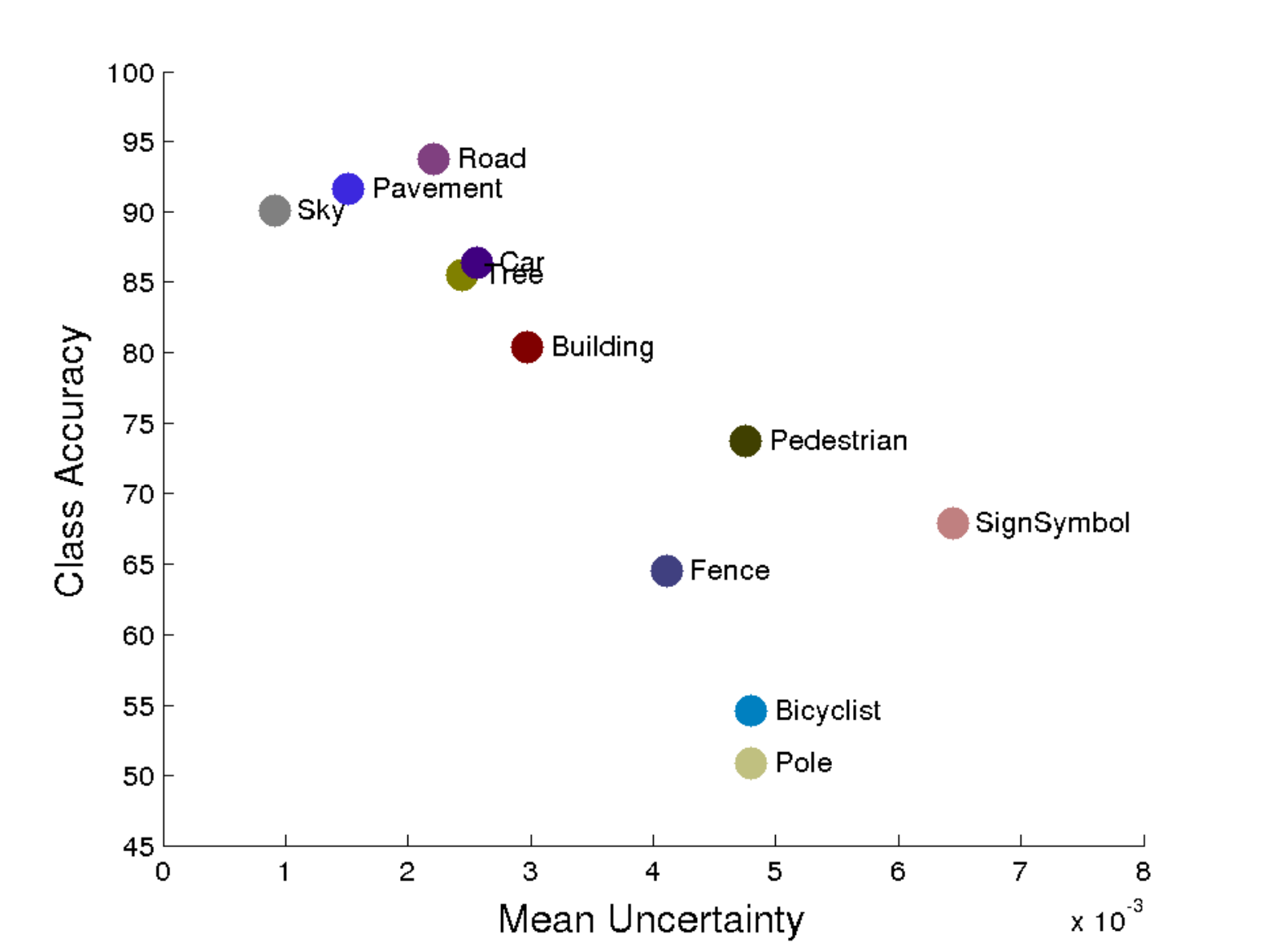}
	\end{center}
	\caption{\textbf{Bayesian SegNet performance compared to mean model uncertainty for each class in CamVid road scene understanding dataset.} This figure shows that there is a strong inverse relationship between class accuracy and model uncertainty. It shows that the classes that Bayesian SegNet performs better at, such as Sky and Road, it is also more confident at. Conversely, for the more difficult classes such as Sign Symbol and Bicyclist, Bayesian SegNet has a much higher model uncertainty.}
	\label{fig:unc_acc}
\end{figure}

\begin{figure}[t]
	\begin{center}
		\includegraphics[width=0.7\linewidth]{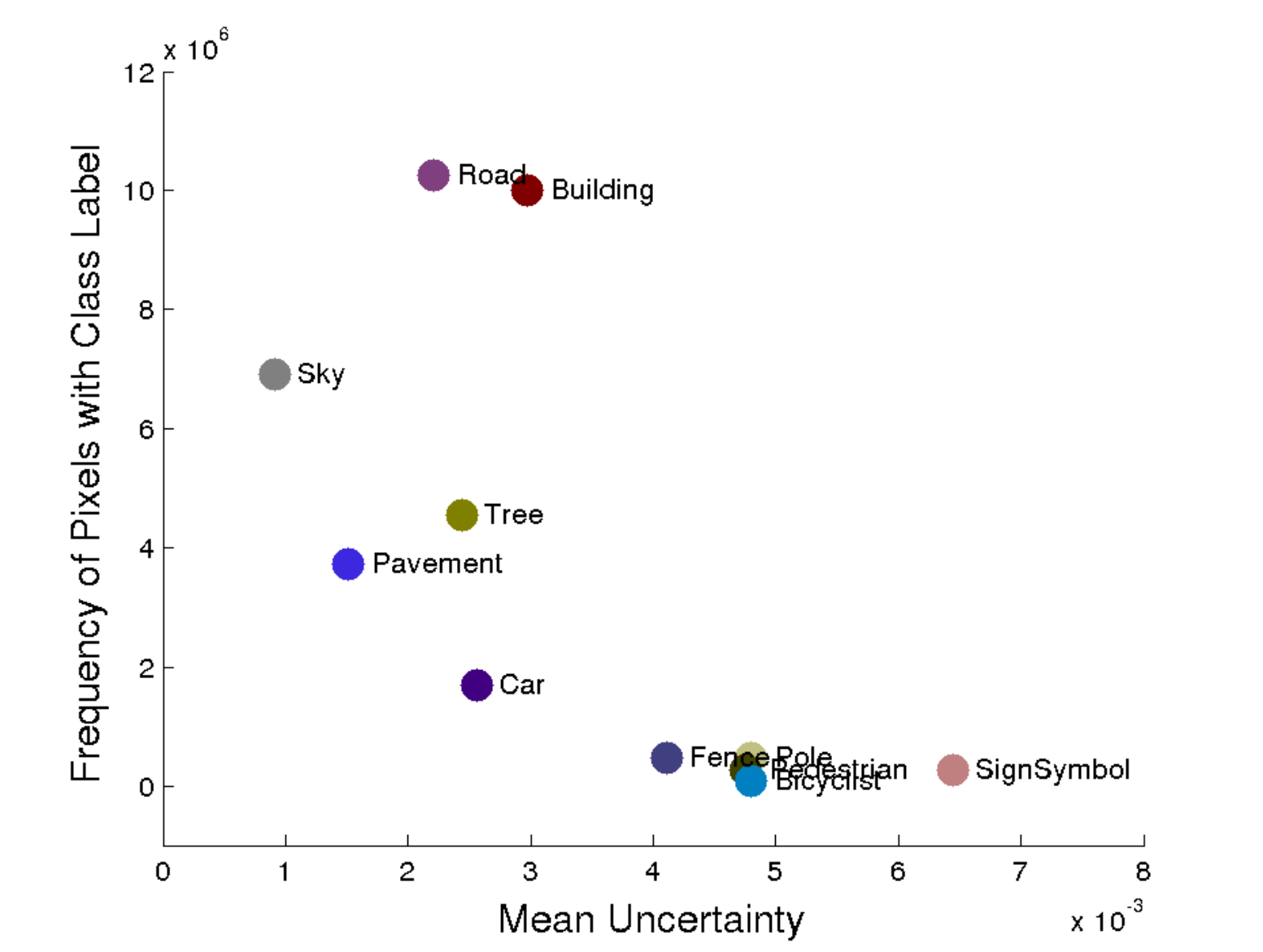}
	\end{center}
	\caption{\textbf{Bayesian SegNet class frequency compared to mean model uncertainty for each class in CamVid road scene understanding dataset.} This figure shows that there is a strong inverse relationship between model uncertainty and the frequency at which a class label appears in the dataset. It shows that the classes that Bayesian SegNet is more confident at are more prevalent in the dataset. Conversely, for the more rare classes such as Sign Symbol and Bicyclist, Bayesian SegNet has a much higher model uncertainty.}
	\label{fig:unc_freq}
\end{figure}

\begin{table*}[t]
\centering
\tabcolsep=3pt
\resizebox{\textwidth}{!}{
	\small{
		\begin{tabular}{c|c|c|c|c|c|c|c|c|c|c|c|c|c|c|c|c|c|c|c|c|c|c|c|c|c|c|c|c|c|c|c|c|c|c|c|c|c}
			& \rotatebox{90}{Wall} & \rotatebox{90}{Floor} & \rotatebox{90}{Cabinet} & \rotatebox{90}{Bed} & \rotatebox{90}{Chair} & \rotatebox{90}{Sofa} & \rotatebox{90}{Table} & \rotatebox{90}{Door} & \rotatebox{90}{Window} & \rotatebox{90}{Bookshelf} & \rotatebox{90}{Picture} & \rotatebox{90}{Counter} & \rotatebox{90}{Blinds} & \rotatebox{90}{Desk} & \rotatebox{90}{Shelves} & \rotatebox{90}{Curtain} & \rotatebox{90}{Dresser} & \rotatebox{90}{Pillow} & \rotatebox{90}{Mirror} & \rotatebox{90}{Floor Mat} & \rotatebox{90}{Clothes} & \rotatebox{90}{Ceiling} & \rotatebox{90}{Books} & \rotatebox{90}{Fridge} & \rotatebox{90}{TV} & \rotatebox{90}{Paper} & \rotatebox{90}{Towel} & \rotatebox{90}{Shower curtain} & \rotatebox{90}{Box} & \rotatebox{90}{Whiteboard} & \rotatebox{90}{Person} & \rotatebox{90}{Night stand} & \rotatebox{90}{Toilet} & \rotatebox{90}{Sink} & \rotatebox{90}{Lamp} & \rotatebox{90}{Bathtub} & \rotatebox{90}{Bag}  \\ \hline \hline
			
			SegNet \cite{badrinarayanan2015segnet} & \rotatebox{90}{\textbf{86.6}} & \rotatebox{90}{\textbf{92.0}} & \rotatebox{90}{52.4} & \rotatebox{90}{\textbf{68.4}} & \rotatebox{90}{\textbf{76.0}} & \rotatebox{90}{54.3} & \rotatebox{90}{59.3} & \rotatebox{90}{37.4} & \rotatebox{90}{53.8} & \rotatebox{90}{29.2} & \rotatebox{90}{49.7} & \rotatebox{90}{32.5} & \rotatebox{90}{31.2} & \rotatebox{90}{17.8} & \rotatebox{90}{5.3} & \rotatebox{90}{53.2} & \rotatebox{90}{28.8} & \rotatebox{90}{36.5} & \rotatebox{90}{29.6} & \rotatebox{90}{0.0} & \rotatebox{90}{14.4} & \rotatebox{90}{67.7} & \rotatebox{90}{32.4} & \rotatebox{90}{10.2} & \rotatebox{90}{18.3} & \rotatebox{90}{19.2} & \rotatebox{90}{11.5} & \rotatebox{90}{0.0} & \rotatebox{90}{8.9} & \rotatebox{90}{38.7} & \rotatebox{90}{4.9} & \rotatebox{90}{22.6} & \rotatebox{90}{55.6} & \rotatebox{90}{52.7} & \rotatebox{90}{27.9} & \rotatebox{90}{29.9} & \rotatebox{90}{8.1}  \\ \hline
			
			Bayesian SegNet & \rotatebox{90}{80.2} & \rotatebox{90}{90.9} & \rotatebox{90}{\textbf{58.9}} & \rotatebox{90}{64.8} & \rotatebox{90}{\textbf{76.0}} & \rotatebox{90}{\textbf{58.6}} & \rotatebox{90}{\textbf{62.6}} & \rotatebox{90}{\textbf{47.7}} & \rotatebox{90}{\textbf{66.4}} & \rotatebox{90}{\textbf{31.2}} & \rotatebox{90}{\textbf{63.6}} & \rotatebox{90}{\textbf{33.8}} & \rotatebox{90}{\textbf{46.7}} & \rotatebox{90}{\textbf{19.7}} & \rotatebox{90}{\textbf{16.2}} & \rotatebox{90}{\textbf{67.0}} & \rotatebox{90}{\textbf{42.3}} & \rotatebox{90}{\textbf{57.1}} & \rotatebox{90}{\textbf{39.1}} & \rotatebox{90}{\textbf{ 0.1}} & \rotatebox{90}{\textbf{24.4}} & \rotatebox{90}{\textbf{84.0}} & \rotatebox{90}{\textbf{48.7}} & \rotatebox{90}{\textbf{21.3}} & \rotatebox{90}{\textbf{49.5}} & \rotatebox{90}{\textbf{30.6}} & \rotatebox{90}{\textbf{18.8}} & \rotatebox{90}{\textbf{ 0.1}} & \rotatebox{90}{\textbf{24.1}} & \rotatebox{90}{\textbf{56.8}} & \rotatebox{90}{\textbf{17.9}} & \rotatebox{90}{\textbf{42.9}} & \rotatebox{90}{\textbf{73.0}} & \rotatebox{90}{\textbf{66.2}} & \rotatebox{90}{\textbf{48.8}} & \rotatebox{90}{\textbf{45.1}} & \rotatebox{90}{\textbf{24.1}}  \\
		\end{tabular}
	}}
	\vspace*{0.1cm}
	\caption{Class accuracy of Bayesian SegNet predictions for the 37 indoor scene classes in the \textbf{SUN RGB-D benchmark dataset} \cite{song2015sun}.}
	\label{SUNRGBDClassavg}
\end{table*}

\textbf{Quantitative observations.} To understand what causes the model to be uncertain, we have plotted the relationship between uncertainty and accuracy in Fig. \ref{fig:unc_acc} and between uncertainty and the frequency of each class in the dataset in Fig. \ref{fig:unc_freq}. Uncertainty is calculated as the mean uncertainty value for each pixel of that class in a test dataset. We observe an inverse relationship between uncertainty and class accuracy or class frequency. This shows that the model is more confident about classes which are easier or occur more often, and less certain about rare and challenging classes.

Additionally, Table \ref{tbl:confidence_percentiles} shows segmentation accuracies for varying levels of confidence. We observe very high levels of accuracy for values of model uncertainty above the 90th percentile across each dataset. This demonstrates that the model's uncertainty is an effective measure of confidence in prediction.

\subsection{Real Time Performance}

Table \ref{tbl:pascal} shows that SegNet and Bayesian SegNet maintains a far lower parameterisation than its competitors. Monte Carlo sampling requires additional inference time, however if model uncertainty is not required, then the weight averaging technique can be used to remove the need for sampling (Fig. \ref{fig:samples} shows the performance drop is modest). Our implementation can run SegNet at 35ms per frame and Bayesian SegNet with 10 Monte Carlo samples at 90ms per frame on Titan X GPU. However inference time will depend on the implementation.

\section{Conclusions}

We have presented Bayesian SegNet, the first probabilistic framework for semantic segmentation using deep learning, which outputs a measure of model uncertainty for each class. We show that the model is uncertain at object boundaries and with difficult and visually ambiguous objects. We quantitatively show Bayesian SegNet produces a reliable measure of model uncertainty and is very effective when modelling smaller datasets. Bayesian SegNet outperforms shallow architectures which use motion and depth cues, and other deep architectures. We obtain the highest performing result on CamVid road scenes and SUN RGB-D indoor scene understanding datasets. We show that the segmentation model can be run in real time on a GPU. For future work we intend to explore how video data can improve our model's scene understanding performance.

{\footnotesize
\bibliographystyle{ieee}
\bibliography{bib}
}

\end{document}